\documentclass[twocolumn]{svjour3}         
\smartqed
\usepackage{times}
\usepackage{epsfig}
\usepackage{graphicx}
\usepackage{amsmath}
\usepackage{amssymb}
\usepackage{caption}

\usepackage{comment}
\usepackage{amsmath,amssymb}
\usepackage{algorithm}
\usepackage{algorithmic}
\usepackage{tabularx}
\usepackage{multirow}
\usepackage{bm}
\usepackage{booktabs}
\usepackage{array}
\usepackage{microtype}
\usepackage{arydshln}
\usepackage[switch]{lineno}

\usepackage{pifont}
\usepackage[usenames,dvipsnames]{xcolor}
\definecolor{dgreen}{rgb}{0.0,0.6,0.0} 
\definecolor{dred}{rgb}{0.6,0.0,0.0} 
\definecolor{alexey}{rgb}{0.7,0,1}
\definecolor{philipcolor}{rgb}{0,0.5,0}
\definecolor{grey}{rgb}{0.6,0.6,0.6}
\definecolor{dblue}{rgb}{0.0,0.0,0.7}

\newcommand{\cmark}{\textcolor{dgreen}{\text{\ding{51}}}}%
\newcommand{\xmark}{\textcolor{dred}{\text{\ding{55}}}}%

\newcommand{\adastereo}{\textit{AdaStereo}}
\newcommand{\adaresnetcorr}{\textit{Ada-ResNetCorr}}
\newcommand{\adapsmnet}{\textit{Ada-PSMNet}}

\usepackage{soul} 

\usepackage[colorlinks,linkcolor=red]{hyperref}

\journalname{IJCV}

\begin{document}

\title{AdaStereo: An Efficient Domain-Adaptive Stereo Matching Approach}

\titlerunning{AdaStereo: An Efficient Domain-Adaptive Stereo Matching Approach}

\author{\small{Xiao Song \and Guorun Yang \and Xinge Zhu \and Hui Zhou \and Yuexin Ma \and Zhe Wang \and Jianping Shi}}

\authorrunning{Xiao Song \emph{et al.}}

\institute{Xiao Song, Hui Zhou, Zhe Wang and Jianping Shi are with the SenseTime Group Limited, email: \{songxiao,zhouhui,wangzhe,shijianping\}@sensetime.com.  Guorun Yang is with the Shenzhen Institutes of Advanced Technology, Chinese Academy of Sciences, email: yangguorun91@gmail.com. Xinge Zhu is with the Chinese University of Hong Kong, email: zhuxinge123@gmail.com. Yuexin Ma is with the ShanghaiTech University, email: mayuexin@shanghaitech.edu.cn.}

\date{Received: 18 February 2021 / Accepted: 11 November 2021}

\maketitle

	\begin{abstract}
	
	 
	 
	 Recently, records on stereo matching benchmarks are constantly broken by end-to-end disparity networks. However, the domain adaptation ability of these deep models is quite limited. Addressing such problem, we present a novel domain-adaptive approach called \textbf{\adastereo} that aims to align multi-level representations for deep stereo matching networks. Compared to previous methods, our AdaStereo realizes a more standard, complete and effective domain adaptation pipeline. Firstly, we propose a non-adversarial progressive color transfer algorithm for input image-level alignment. Secondly, we design an efficient parameter-free cost normalization layer for internal feature-level alignment. Lastly, a highly related auxiliary task, self-supervised occlusion-aware reconstruction is presented to narrow the gaps in output space. We perform intensive ablation studies and break-down comparisons to validate the effectiveness of each proposed module. With no extra inference overhead and only a slight increase in training complexity, our AdaStereo models achieve state-of-the-art cross-domain performance on multiple benchmarks, including KITTI, Middlebury, ETH3D and DrivingStereo, even outperforming some state-of-the-art disparity networks finetuned with target-domain ground-truths. Moreover, based on two additional evaluation metrics, the superiority of our domain-adaptive stereo matching pipeline is further uncovered from more perspectives. Finally, we demonstrate that our method is robust to various domain adaptation settings, and can be easily integrated into quick adaptation application scenarios and real-world deployments.
    \keywords{Domain adaptation \and Stereo matching \and Non-adversarial progressive color transfer \and Cost normalization \and Self-supervised occlusion-aware reconstruction}
	\end{abstract}

\section{Introduction}
\label{sec:intro}
	
	Stereo matching is a fundamental problem in computer vision. The task aims to find corresponding pixels in a stereo pair, and the distance between corresponding pixels is known as disparity \cite{hartley2003multiple}. Based on the epipolar geometry, stereo matching enables stable depth perception from estimated disparity, hence it has been applied to further applications such as scene understanding~\cite{franke2000real,miclea2019real,zhang2010semantic}, object detection~\cite{chen20153d,li2018stereo,li2019stereo,fang2018putting}, visual odometry~\cite{wang2017stereo,zhu2017image}, and SLAM~\cite{engel2015large,gomez2019pl}.
	
	\begin{figure*}[!t]
		\centering
		\includegraphics[width=0.90\textwidth]{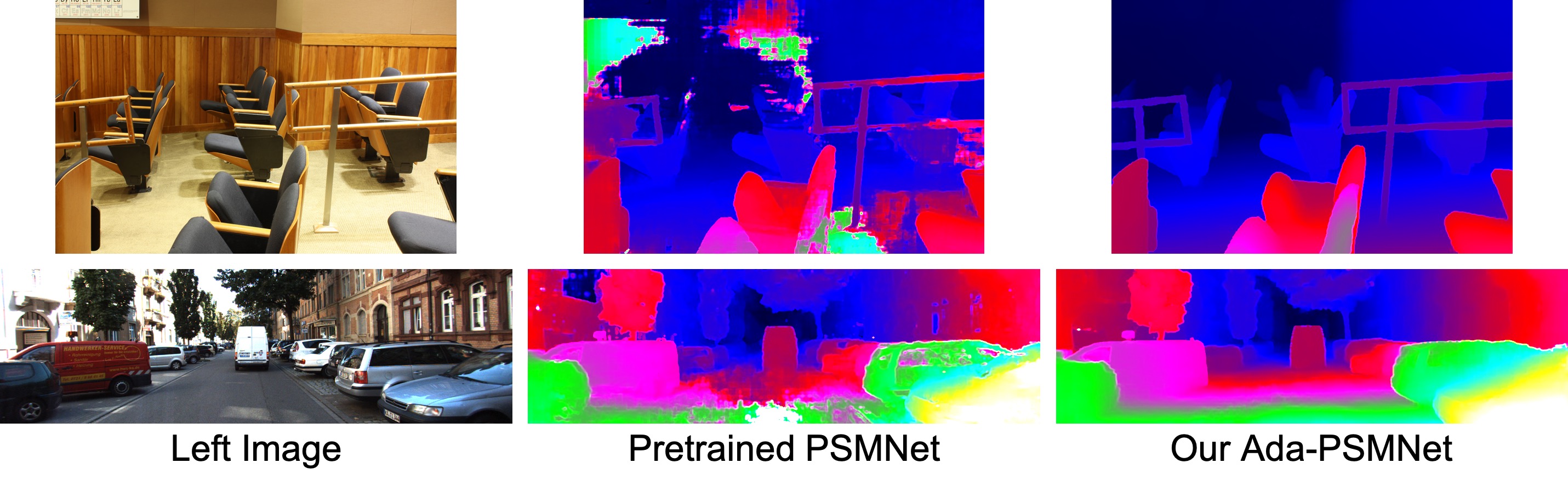}
		\caption{
			\textbf{Overview examples.} Top-down: Middlebury~\cite{Scharstein2014High} and KITTI~\cite{Menze2015CVPR} examples. Left-right: left stereo image, disparity map predicted by the SceneFlow-pretrained PSMNet~\cite{chang2018pyramid}, and disparity map predicted by our Ada-PSMNet.
		}
		\vspace{-12pt}
		\label{fig:overview_examples}
	\end{figure*}
	
 	Classical stereo matching methods generally use hand-craft features and pre-defined windows to compute and aggregate matching costs. Due to the weak feature representations, most classical methods have problems in the areas with no textures, perspective distortions, and illumination changes. Since deep neural networks make great progress in the image classification task~\cite{krizhevsky2012imagenet,simonyan2014very,szegedy2015going,he2016deep}, Zbontar and Lecun \cite{zbontar2015computing} first introduced the deep network to extract image features and compute matching costs, which outperforms traditional methods by a significant margin. In order to further improve the accuracy and efficiency of stereo matching, recent methods typically adopt fully convolutional networks~\cite{long2015fully} to directly regress disparity maps. Utilizing the massive synthetic data~\cite{mayer2016large} rendered by graphics tools for pretraining and real-world ground-truths for finetuning, these methods have achieved state-of-the-art performance on public stereo matching   benchmarks~\cite{Geiger2012CVPR,Menze2015CVPR,Scharstein2014High}. However, the performance of these methods adapted from synthetic data to real-world scenes is quite limited, which is known as the domain adaptation problem. Here, we select the PSMNet~\cite{chang2018pyramid} to illustrate this problem. As shown in Fig.~\ref{fig:overview_examples}, the PSMNet pretrained on the SceneFlow dataset~\cite{mayer2016large} works well on the synthetic image pairs, but fails to produce equally good results on the Middlebury~\cite{Scharstein2014High} and KITTI~\cite{Menze2015CVPR} datasets, in which disparity predictions are incorrect on the roads, vehicle windows, and indoor walls. Therefore instead of designing powerful models for higher accuracy on specific datasets, how to obtain effective domain-adaptive stereo matching networks is more desirable now.


    In this work, we aim at the important but less explored problem of domain adaptation in stereo matching. Since there is a great number of synthetic data but only a small amount of realistic data with ground-truths, hence we focus on domain gaps between synthetic and real domains. Feeding two pairs of images under distinct domains into the stereo matching model, we analyze the differences at various levels, as shown in Fig.~\ref{fig:feature_visualization}. At the input image level, color and brightness are the obvious gaps. We then plot a histogram for the values in the cost volume, which is a crucial internal representation for stereo matching, and significant differences can be found in the distribution of cost values.  Moreover, geometries of the output disparity map are inconsistent as well, \emph{e.g.} the shape of objects, and the relative position between foreground and background. In order to bridge the domain gaps at these levels, \emph{i.e.} input image, internal cost volume, and output disparity, we propose a standard and complete domain adaptation pipeline for stereo matching named \textbf{\adastereo}, in which three particular modules are presented: 
    
	
	
	\begin{itemize}
	    \item[\small{$\bullet$}] For input image-level alignment, an effective algorithm, \textbf{non-adversarial progressive color transfer}, is proposed to align input color space between source and target domains during training. It is the first attempt that adopts non-adversarial style transfer method to align input-level inconsistency in the stereo domain adaptation, avoiding side-effects of harmful geometrical distortions caused by GAN-based methods \cite{zhu2017unpaired}. Furthermore, the proposed progressive update scheme enables capturing representative target-domain color styles during adaptation. 
	    
	    
	
	    \item[\small{$\bullet$}] For internal feature-level alignment, a \textbf{cost normalization} layer is designed to align the distribution of cost volumes. Oriented to the stereo matching task, we propose two normalization operations: (i) channel normalization reduces the inconsistency in scaling of each feature channel; and (ii) pixel normalization further regulates the norm distribution of pixel-wise feature vector for binocular matching. Embedded these two operations in one layer, the inter-domain gaps in matching costs are effectively narrowed. It is worth noting that our cost normalization layer is parameter-free and adopted only once in the network compared to previous general normalization layers (\emph{e.g.} IN \cite{ulyanov2016instance}, DN \cite{zhang2020domain}).
	    

	    
	    
	    \item[\small{$\bullet$}] For output-space alignment, we conduct self-supervised learning through a highly related auxiliary task, \textbf{self-supervised occlusion-aware reconstruction}, which is the first proposed auxiliary task for stereo domain adaptation. Concretely, a self-supervised module is attached upon the main disparity network to perform image reconstructions on the target domain. To address the ill-posed occlusion problem in reconstruction, we also design a domain-collaborative learning scheme for occlusion mask predictions. Through occlusion-aware stereo reconstruction, more informative geometries from target scenes are involved in model training, so as to benefit the disparity predictions across domains.
	    
	    
    \end{itemize}

	Based on the proposed pipeline, we conduct domain adaptation from synthetic data to real-world scenes. In order to validate the effectiveness of each module, ablation studies are performed on diverse real-world datasets, including Middlebury~\cite{Scharstein2014High}, ETH3D~\cite{schoeps2017cvpr}, KITTI~\cite{Geiger2012CVPR,Menze2015CVPR} and DrivingStereo~\cite{yang2019drivingstereo}. Specifically, for the image-level alignment, we compare our progressive color transfer method with other GAN-based methods. The cost normalization is also compared with those general normalization techniques. Moreover, we experiment different loss combinations to demonstrate the effectiveness of our self-supervised occlusion-aware loss. Eventually, our domain-adaptive models outperform other traditional / domain generalization / domain adaptation methods and even finetuned disparity networks on multiple public benchmarks. As shown in Fig.~\ref{fig:overview_examples}, our Ada-PSMNet pretrained on  the synthetic dataset performs well on both indoor and outdoor scenes. Main contributions are summarized below:
	
	\begin{itemize}
		\item[\small{$\bullet$}]
		We locate the domain-adaptive problem and investigate domain gaps for deep stereo matching networks. 
		\item[\small{$\bullet$}]
		We propose a novel domain adaptation pipeline, including three modules to narrow the gaps at input image-level, internal feature-level and output space.
		\item[\small{$\bullet$}]
		Our domain-adaptive models outperform other domain-invariant methods, and even finetuned disparity networks on multiple stereo matching benchmarks.
		
		
	\end{itemize}
	
 	 \textbf{Differences with our conference paper \cite{song2021adastereo}.} The extended contents are mainly in the related work, method analyses and descriptions, and abundant extended experiments (more than $3$ new figures, and extra $11$ tables). Specifically, more insightful analyses are included to illustrate the motivations and advantages of each proposed module in our domain-adaptive stereo matching pipeline. In addition, important instructions are supplemented to clarify the structure of self-supervised occlusion-aware reconstruction module, following by  corresponding training details. To further reveal the effectiveness and generalization capability of our method, we conduct more experiments, including abundant extended ablation studies, additional cross-domain / benchmark performance comparisons on the DrivingStereo test set, as well as more verification and novel applications (\emph{i.e.} new evaluation metrics, verification on top of more advanced stereo networks, real-world source domain applications, semi-supervised applications, and quick adaptations).

\section{Related Work}
	\label{sec:related_work}
	
	\subsection{Stereo Matching}
	
	The classical pipeline of stereo matching usually follows the four-step pipeline~\cite{scharstein2002taxonomy}: matching cost computation, matching cost aggregation, disparity computation, and disparity refinement. According to the optimization mode, the methods can be divided into local matching methods and global matching methods. The local matching methods~\cite{okutomi1993multiple,zabih1994non,kang1995multibaseline,black1996unification,di2005zncc,mattoccia2008fast} focus on the computation and aggregation of matching costs to find matching pixels within a certain range. The global methods transform the stereo matching into an energy-minimization problem, followed by different solutions, \emph{e.g.} dynamic programming~\cite{ohta1985stereo}, belief propagation~\cite{sun2003stereo}, or graph cut~\cite{roy1998maximum}. Generally, local methods have high speed but low accuracy, while global methods achieve higher accuracy but suffer slower speed. Furthermore, both local methods and global methods are difficult to handle the areas with no textures, perspective distortions, and illumination changes. 
	
	With the breakthrough of deep neural networks in the image classification task~\cite{krizhevsky2012imagenet,simonyan2014very,szegedy2015going,he2016deep}, such structure also achieved great success in the stereo matching task by representing image patches with deep features and computing patch-wise similarity scores \cite{zbontar2015computing,luo2016efficient,shaked2016improved,chen2015deep}. Benefited from stronger feature representations, these methods achieve much higher accuracy than traditional methods. However, due to the limited receptive field, the performance of these methods is unsatisfactory especially in ill-posed regions, and the processing is  still time-consuming because of the remained hand-engineered post-processing steps.

    Inspired from the semantic segmentation task~\cite{long2015fully,chen2015semantic}, recent methods~\cite{mayer2016large,kendall2017end,saikia2019autodispnet,cheng2020hierarchical} adopt fully-convolutional networks to regress disparity maps, which further improves the accuracy and efficiency of stereo matching. Concretely, current end-to-end stereo matching networks can be roughly categorized into two types: correlation-based $2$-D stereo networks and cost-volume based $3$-D stereo networks. For the first category, Mayer \emph{et al.} \cite{mayer2016large} proposed the first end-to-end disparity network DispNetC, in which warping between features of two views is conducted for matching cost calculation, and per-pixel disparity is directly regressed without any post-processing steps. Based on color or feature correlations, more advanced methods were proposed, including CRL \cite{pang2017cascade}, iResNet \cite{liang2017learning}, HD$^{3}$ \cite{yin2019hierarchical}, SegStereo \cite{yang2018SegStereo}, EdgeStereo \cite{song2018stereo,song2020edgestereo}, \emph{etc}. For the second category, $3$-D convolutional neural networks show the advantages in regularizing cost volume for disparity estimation \cite{yang2019hierarchical,wu2019semantic,zhang2020adaptive,xu2020aanet,gu2020cascade}. GC-Net \cite{kendall2017end} first introduced the $4$-D cost volume without collapsing the feature dimension. PSMNet \cite{chang2018pyramid} further applied a pyramid feature extraction module and stacked $3$-D hourglass blocks to improve the accuracy of disparity prediction. Recently, more advanced methods were proposed to further optimize the cost volume regularization, including GwcNet \cite{guo2019group}, EMCUA \cite{nie2019multi}, CSPN \cite{cheng2019learning}, GANet \cite{zhang2019ga}, \emph{etc}. Our proposed domain adaptation pipeline for stereo matching can be easily applied to both $2$-D and $3$-D stereo networks.
    
    
    
	\subsection{Domain Adaptation}
     
     Prior works on domain adaptation can be roughly divided into three categories. The first category mainly covers the conventional methods including discrepancy measures such as MMD \cite{MMD,MMD2} and CMD \cite{zellinger2017central}, geodesic flow kernel~\cite{GFK}, sub-space alignment~\cite{SA}, asymmetric metric learning~\cite{sml}, \emph{etc}. The general idea of the second category is to align source and target domains at different representation levels, including: (1) input image-level alignment \cite{bousmalis2017unsupervised,hoffman2018cycada} using image-to-image translation methods such as CycleGAN \cite{zhu2017unpaired}, or statistics matching \cite{abramov2020keep}; (2) internal feature-level alignment based on feature-level domain adversarial learning \cite{tzeng2017adversarial,long2018conditional,zhao2019geometry}; and (3) output-space alignment \cite{adapt,vu2019advent,lopez2020desc} typically by an adversarial module. For the third category, self-supervised learning based domain adaptation methods \cite{ghifary2016deep} achieve great progress, in which simple auxiliary tasks generated automatically from unlabeled data are utilized to train feature representations, such as rotation prediction \cite{gidaris2018unsupervised}, flip prediction \cite{xu2019self}, patch location prediction \cite{xu2019self}, \emph{etc}. In this paper, we explicitly implement domain alignments at input level and internal feature level, while incorporating self-supervised learning into output-space alignment through a specifically designed auxiliary task.
     
     In addition, the existing domain adaptation methods are mainly designed for high-level tasks such as classification, semantic segmentation and object detection. For semantic segmentation, the pioneering work is~\cite{FCNwild}, which combined global and local alignment methods with domain adversarial training. In~\cite{cda,pena,adapt,chen2018road,gong2019dlow,zheng2018t2net}, authors performed output-space adaptation at feature level by an adversarial module. For object detection, strong-weak alignment~\cite{saito2019strong} and local region alignment~\cite{zhu2019adapting} were applied to tackle the difficulty of domain adaptation. However, much less attention has been paid to domain adaptation for low-level tasks. There are several works investigating the domain-adaptive depth estimation task, including geometry-aware alignment~\cite{zhao2019geometry}, semantic-level consistency~\cite{lopez2020desc} and image-level translation~\cite{zheng2018t2net,atapour2018real}.
     

	
   \subsection{Domain-adaptive Stereo Matching}
   
   
    
     Although records on public benchmarks are constantly broken, few attention has been paid to the domain adaptation ability of deep stereo networks. For stereo matching domain adaptation, Pang \emph{et al.} \cite{pang2018zoom} proposed a semi-supervised method utilizing the scale information. Guo \emph{et al.} \cite{guo2018learning} presented a cross-domain method using knowledge distillation. MAD-Net \cite{tonioni2019real} was designed to adapt a compact stereo model online. Recently, StereoGAN~\cite{liu2020stereogan} utilized CycleGAN~\cite{zhu2017unpaired} to bridge domain gaps by joint optimizations of image style transfer and stereo matching. However, no standard and complete domain adaptation pipeline was implemented in these methods, and their adaptation performance is quite limited. Contrarily, we propose a more complete pipeline for deep stereo models following the standard domain adaptation methodology, in which alignments across domains are conducted at multiple representation levels thereby remarkable adaptation performance is achieved. In addition, we do not conduct any adversarial learning hence the training stability and semantic invariance are guaranteed.
     


\section{Method}
	\label{sec:method}

    
    In this section, we first describe the problem of domain-adaptive stereo matching. Then we introduce the motivation and give an overview of our domain adaptation pipeline. After that, we detail the main components in the pipeline, \emph{i.e.} non-adversarial progressive color transfer, cost normalization and self-supervised occlusion-aware reconstruction.
	
	\subsection{Problem Description}
	\label{sec:formulation}
	
	In this paper, we focus on the \textbf{domain adaptation} problem for stereo matching. Differing from domain generalization where a method needs to perform well on unseen scenes, domain adaptation enables methods to use target-domain images without target-domain ground-truths during training. Specifically for stereo matching, since there is a large amount of synthetic data~\cite{mayer2016large} but only a small number of realistic data with ground-truths~\cite{Menze2015CVPR,Scharstein2014High,schoeps2017cvpr}, the problem can be further limited to the adaptation from virtual to real-world scenes. Given stereo image pairs $({I_s^l}, {I_s^r})$ and $({I_t^l}, {I_t^r})$ on source and target domains, and the ground-truth disparity map $\hat{d_{s}^{l}}$ on the source domain, we train the model to predict the disparity map $d_t^l$ on the target domain.
	
	\begin{figure}[tb]
		\centering
		\includegraphics[width=1.0\linewidth]{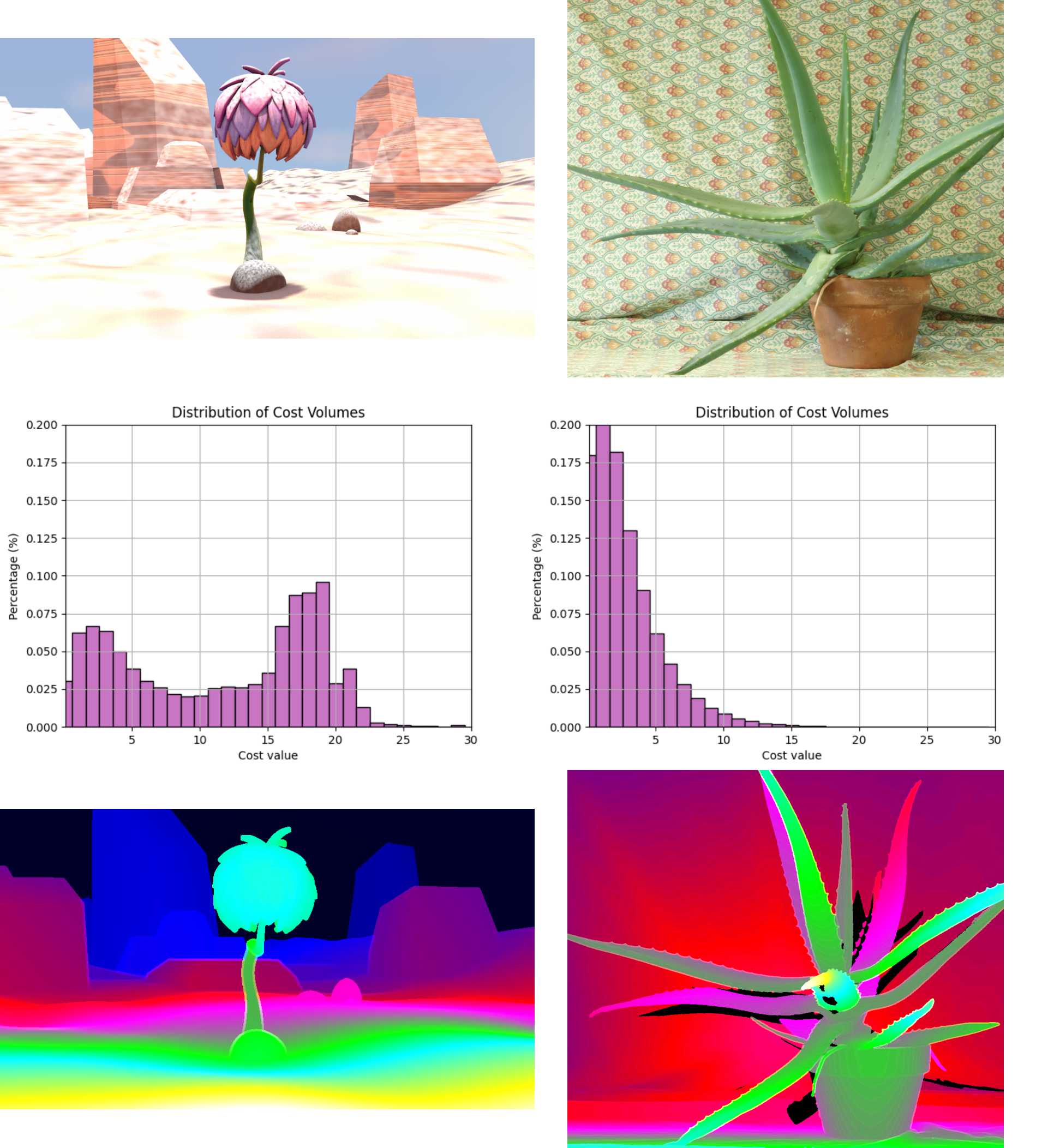}
		\caption{
			\textbf{Comparisons of input images, internal representations and output disparities between synthetic and real-world datasets.} Left-right: SceneFlow~\cite{mayer2016large} and Middlebury~\cite{Scharstein2014High} examples. Top-down: input image, internal cost volume and output disparity map. Disparity maps of two domains are rendered by the same color  map.
		}
		\label{fig:feature_visualization}
		\vspace{-10pt}
	\end{figure}
	
	\subsection{Motivation}
	\label{sec:motivation}
		
	
	\begin{figure*}[t]
	\centering
	\includegraphics[width=0.99\textwidth]{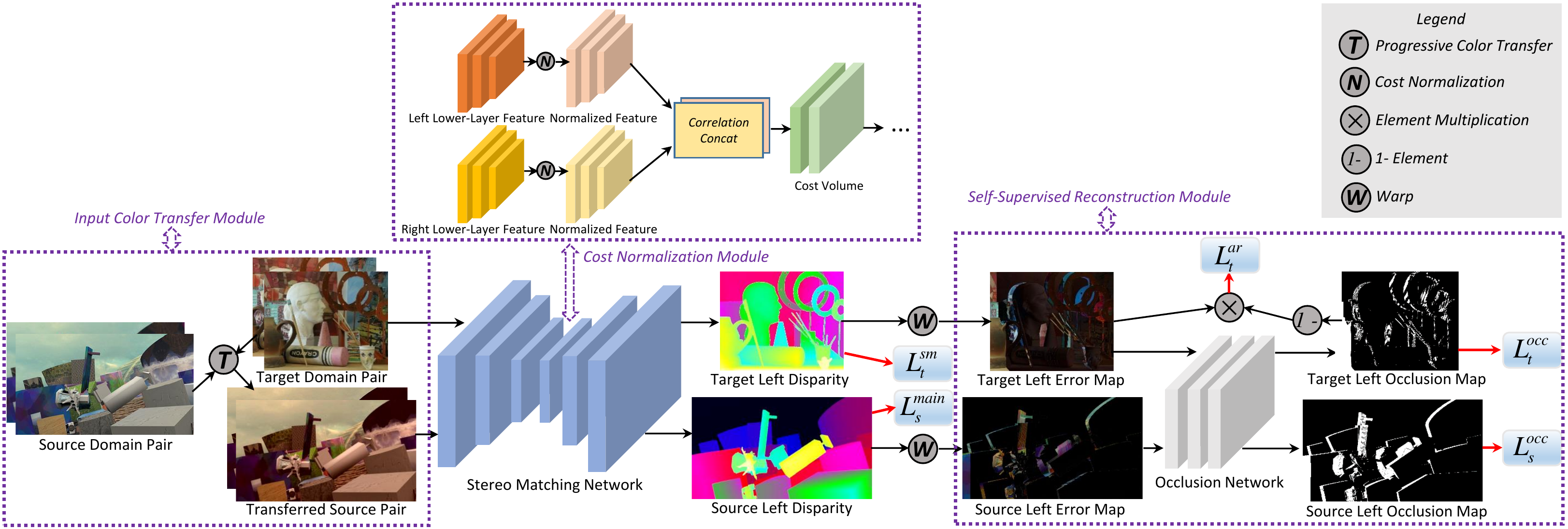}
	\caption{
		\textbf{The training diagram of our~\adastereo}, with the adaptation from SceneFlow to Middlebury as an example. Color transfer and self-supervised occlusion-aware reconstruction modules are only adopted during training. ($L_{s}^{main}, L_{s}^{occ}$, $L_{t}^{ar}$, $L_{t}^{occ}$,  $L_{t}^{sm}$) are the five training loss terms specified in Eq.~\ref{loss_t}.
	}
	\label{fig:training_diagram} 
	\vspace{-10pt}
	\end{figure*}
	 
	As shown in Fig.~\ref{fig:feature_visualization}, two images from SceneFlow~\cite{mayer2016large} and Middlebury~\cite{Scharstein2014High} datasets are selected to describe inherent inconsistencies between two domains. (i) These two images own observable differences in their color and brightness. Moreover, according to the statistics on \emph{whole datasets}, the mean values of RGB channels are $(107, 102, 92)$ in SceneFlow and $(148, 132, 102)$ in Middlebury, in which significant color variances are found between synthetic and realistic domains. (ii) For cost volumes computed from the 1D-correlation layer ~\cite{mayer2016large}, we analyse the distribution of matching cost values. In the second row of Fig.~\ref{fig:feature_visualization}, the cost values, ranging from $0$ to $30$, are divided into $30$ bins. We plot the histogram to depict the proportion of matching cost values in each interval and find the distributions between two domains are inconsistent. Specifically, the cost values of SceneFlow scatter between $0{\sim}25$, while the cost values of Middlebury mainly distribute between $0{\sim}10$. Since the cost values are the products of binocular features, the distributions show the differences of binocular features for cost volume construction across domains. (iii) Although the two sampled images have similar plants as foreground, the generated disparity maps still vary in the scene geometries. Both the foreground objects and background screens have quite different disparities. Therefore, we conclude that the inherent differences across domains for stereo matching lie in the image color at input level, cost volume at feature level, and disparity map at output level. 
	
	
	
	Correspondingly, to solve the domain adaptation problem of stereo matching, we propose the progressive color transfer, cost normalization, and self-supervised reconstruction to handle the domain gaps in three levels respectively. The former two strategies are presented to directly narrow the differences in color space and cost volume distribution. The latter reconstruction is appended as an auxiliary task to impose extra supervision on the estimated disparity map for target domain and finally benefits the adaptation ability.
	

	\subsection{Framework Overview}
	
    
    Fig.~\ref{fig:training_diagram} depicts the training pipeline of our stereo domain adaptation framework, in which three major modules are involved. For input, a randomly selected target-domain pair and a randomly selected source-domain pair which is adapted to target-domain color styles through our \emph{progressive color transfer} algorithm, are simultaneously fed into a shared-weight disparity network with our \emph{cost normalization} layer. The source-domain branch is under the supervision of the  given ground-truth disparity map, while the target-domain branch is regulated by the proposed auxiliary task: \emph{self-supervised occlusion-aware reconstruction}.

	
	

	\subsection{Non-adversarial Progressive Color Transfer}
	
	
	
	As mentioned in Sec.~\ref{sec:motivation}, the color difference plays a major role in the input image-level inconsistency across domains. Hence, we present an effective and stable algorithm for color transfer from source domain to target domain in a non-adversarial manner. During training, given a source-domain image $I_s$ and a target-domain image $I_t$, the algorithm outputs a transferred source-domain image $I_{s\rightarrow t}$, which preserves the content of $I_s$ and owns the target-domain color style.  In Alg.~\ref{alg:color_transfer}, the transfer is performed in the $LAB$ color space. ${T}_{RGB{\rightarrow}LAB}(.)$ and ${T}_{LAB{\rightarrow}RGB}(.)$ denote the color space transformations. Under the $LAB$ space, the mean value $\mu$ and standard deviation $\sigma$ of each channel are first computed by $S(\cdot)$. Then, each channel in the source-domain $LAB$ image $\tilde{I_s}$ is subtracted by its mean ${\mu}_{s}$ and multiplied by the standard deviation ratio $\lambda$. Finally, the transferred image $I_{s\rightarrow t}$ is obtained through the addition of the progressively updated ${\mu}_{t}$ and color space conversion. In the training pipeline of ~\adastereo, two images in a source-domain pair are simultaneously transferred with the same ${\mu}_{t}$ and $ {\sigma}_{t}$.
	
	\begin{algorithm}
		\caption{\textbf{Progressive Color Transfer}}
		\label{alg:color_transfer}
		\begin{algorithmic}[1]
			\REQUIRE {Source-domain dataset $D_s$, target-domain dataset $D_t$, ${\mu}_{t}=0$, ${\sigma}_{t}=0$}
			\STATE Randomly shuffle $D_s$ and $D_t$
			\FOR{$i \in [0, len(D_s))$}
			\STATE Select $I_s \in D_s$, $I_t \in D_t$
			\STATE $\tilde{I_s}{~\Leftarrow~}{T}_{RGB{\rightarrow}LAB}(I_s)$, \space\space $\tilde{I_t}{~\Leftarrow~}{T}_{RGB{\rightarrow}LAB}(I_t)$
			\STATE $({\mu}_s, {\sigma}_s){~\Leftarrow~}{S{(\tilde{I_s})}}$,
			\space\space
			$({{\mu}_{t}}^{i}, {{\sigma}_{t}}^{i}){~\Leftarrow~}{S{(\tilde{I_t})}}$
			\STATE
			${\mu}_{t}{~\Leftarrow~} (1-\gamma)*{\mu}_{t}+ \gamma*{{\mu}_{t}}^{i}$
			\STATE
			$ {\sigma}_{t}{~\Leftarrow~}{(1-\gamma)*{\sigma}_{t}+\gamma*{{\sigma}_{t}}^{i}}$
			\STATE
			$\tilde{I_s}{~\Leftarrow~}{\tilde{I_s}-{\mu}_s} $, ${\lambda}{~\Leftarrow~}{{\sigma}_t / {\sigma}_s}$
			\STATE $\tilde{I_{s\rightarrow t}}{~\Leftarrow~}{\lambda * \tilde{I_s}+ {\mu}_t}$
			\STATE ${I_{s\rightarrow t}}{~\Leftarrow~}{T}_{LAB{\rightarrow}RGB}(\tilde{I_{s\rightarrow t}}) $
			\ENDFOR
		\end{algorithmic}
	\end{algorithm}
	
	Compared with Reinhard's color transfer method~\cite{reinhard2001color}, the main contribution of our algorithm is the proposed progressive update scheme that proves to be more beneficial for domain adaptation. Considering color inconsistencies might exist across different images in the same target-domain dataset while the previous method ~\cite{reinhard2001color} only enables one-to-one transformation, the transferred source-domain images can not capture the meaningful color styles that are representative for the whole target-domain dataset. The progressive update scheme is proposed to address such problem. To be specific, target-domain ${\mu}_{t}$ and ${\sigma}_{t}$ are progressively re-weighted by current inputs $({{\mu}_{t}}^{i}, {{\sigma}_{t}}^{i} )$ and historical records $({\mu}_{t}, {\sigma}_{t})$ with a momentum $\gamma$, simultaneously ensuring the representativeness of target-domain color styles and the diversity of training samples transferring from the source domain to the target domain during training. More evidences of its effectiveness are provided in experiments.

	In a larger sense, we are the first to use a non-adversarial style transfer method to align the input-level inconsistency for stereo domain adaptation. Unlike GAN-based style transfer networks~\cite{li2017watergan,zhu2017unpaired} that cause side-effects of harmful geometrical distortions for the low-level stereo matching task, our method is more training-efficient meanwhile ensuring the training stability and semantic invariance, which can be easily embedded into the training framework of domain-adaptive stereo matching. Experimental results further validate its superiority over other adversarial transfer methods.

	\subsection{Cost Normalization}
    
    Cost volume is one of the most important internal feature-level representation in a deep stereo network, encoding all necessary information for succeeding disparity regression.
    Hence for domain-adaptive stereo matching, an intuitive way is to directly narrow the deviations in cost volume distributions across different domains.
    Based on the analyses in Sec.~\ref{sec:motivation}, the cross-domain  differences of cost volumes mainly originate from the binocular features for cost volume construction.  Correspondingly, we design a cost normalization layer to regularize the left and right features, which is compatible with all cost volume building patterns (\emph{e.g.} correlation or concatenation) in stereo matching, as shown in Fig.~\ref{fig:training_diagram}.
    
	
    Before cost volume construction, the left lower-layer feature $\mathcal{F^{L}}$ and right feature $\mathcal{F^{R}}$ with the same size of $N \times C \times H \times W$ ($N$: batch size, $C$: channel, $H$: spatial height, $W$: spatial width), are both successively regularized by two proposed normalization operations: channel normalization and pixel normalization. Specifically, the channel normalization is applied across all spatial dimensions ($H, W$) per channel per sample
    individually, which is defined as:
    \begin{equation}
    \\ \\ \, \mathcal{F}_{n,c,h,w} = \frac{\mathcal{F}_{n,c,h,w}}{\sqrt{\sum_{h=0}^{H-1}{\sum_{w=0}^{W-1}{||\mathcal{F}_{n,c,h,w}||^{2}}}+\varepsilon}}
    \end{equation}
    where $\mathcal{F}$ denotes the lower-layer feature, ${h}$ and ${w}$ denote the spatial position, $c$ denotes the channel, and $n$ denotes the batch index. After the channel normalization, the pixel normalization is further applied across all channels per spatial position per sample individually, which is defined as:
    \begin{equation}
    \\ \\ \mathcal{F}_{n,c,h,w} = \frac{\mathcal{F}_{n,c,h,w}}{\sqrt{\sum_{c=0}^{C-1}{||\mathcal{F}_{n,c,h,w}||^{2}}+\varepsilon}}
    \end{equation}
    Through channel normalization which reduces the inconsistency in norm and scaling of each feature channel, and pixel normalization which further regulates the norm distributions of pixel-wise feature vectors for binocular matching, inter-domain gaps in generated matching costs due to varied image contents and geometries are greatly reduced.
    
    In a nutshell, our \emph{parameter-free} cost normalization layer is indeed a  normalization layer designed specifically for stereo domain adaptation, which is adopted only once before cost volume construction. On the contrary, previous normalization layers (\emph{e.g.} BIN \cite{nam2018batch}, IN \cite{ulyanov2016instance}, CN \cite{dai2019channel} (exactly equivalent to IN) and DN \cite{zhang2020domain}) are general normalization approaches, which contain learnable parameters and are repeatedly adopted in the network's feature extractor. Hence regulations on cost volume from those general normalization layers are not direct and effective enough, requiring extra implicit learning process. Moreover, our cost normalization layer does not use zero-centralization to prevent extra disturbances in matching cost distributions. Experiments further validate its superiority over general normalization layers.
    	
    

    \subsection{Self-supervised Occlusion-aware Reconstruction}
    
    
    Self-supervised auxiliary tasks are demonstrated to be beneficial for aligning domains in high-level tasks \cite{gidaris2018unsupervised,xu2019self}. However, such problem has not been explored for the low-level stereo matching task. Hence in this sub-section, we propose an effective auxiliary task for stereo domain adaptation: self-supervised occlusion-aware reconstruction. As shown in Fig.~\ref{fig:training_diagram}, a self-supervised module is attached upon the main disparity network to perform image reconstructions on the target domain. To address the ill-posed occlusion problem in reconstruction, we design a domain-collaborative occlusion mask learning scheme. Through occlusion-aware stereo reconstruction, more informative geometries from target-domain scenes are involved in training.

\begin{figure}[tb]
		\centering
		\includegraphics[width=1.0\linewidth]{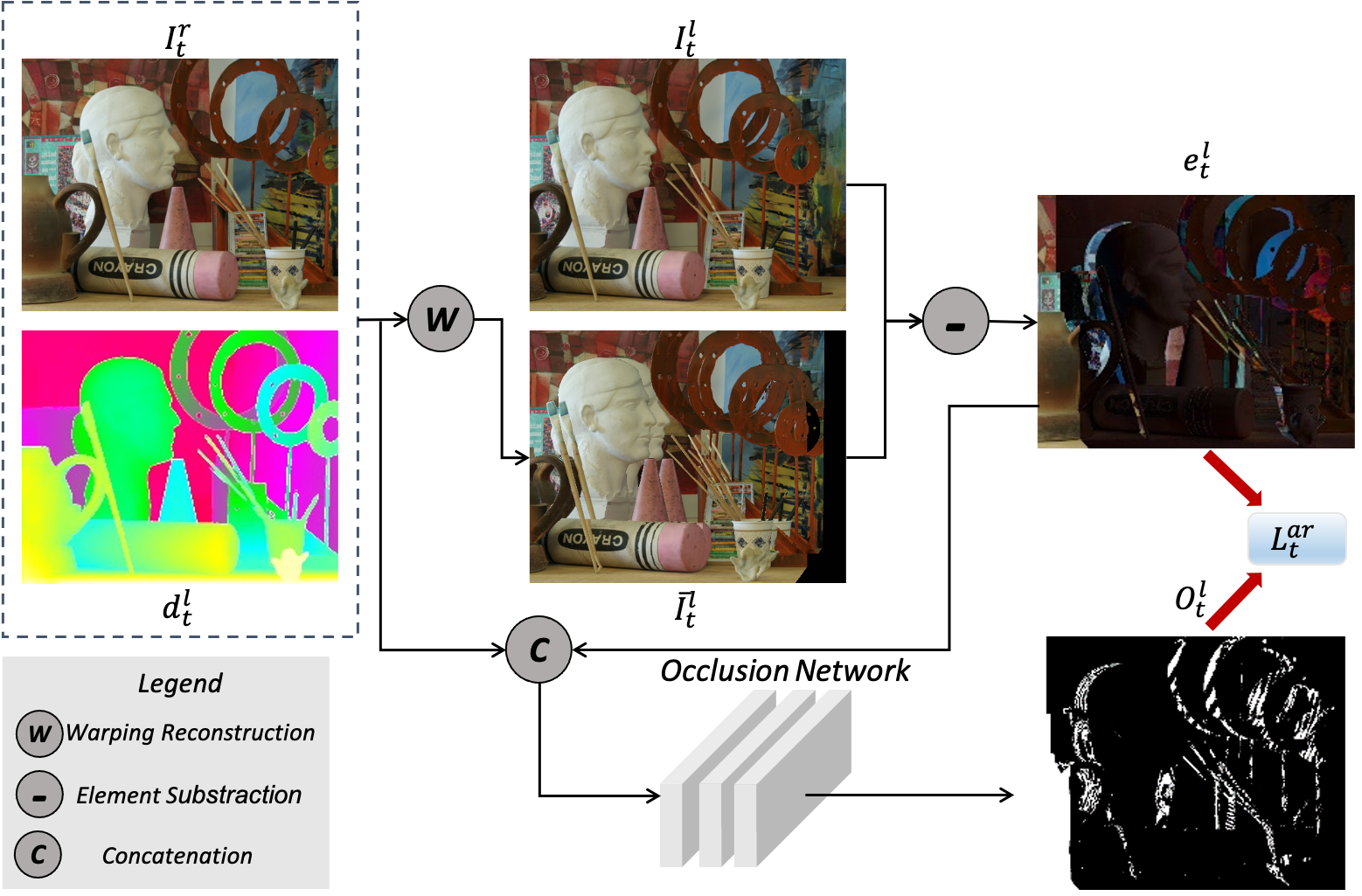}
		\caption{
			The training details of the self-supervised occlusion-aware reconstruction module on the target domain.
		}
		\label{fig-self_supervised}
	\end{figure}

    During the self-supervised learning, stereo reconstruction is firstly measured by the difference between the input target-domain left image $I_{t}^{l}$ and the corresponding warped image $\overline{I_{t}^{l}}$ (based on the right image $I_{t}^{r}$ and the produced disparity map $d_{t}^{l}$). Then a small fully-convolutional occlusion prediction network takes the concatenation of $d_{t}^{l}$, $I_{t}^{r}$ and the pixel-wise error map $e_{t}^{l}=|I_{t}^{l}-\overline{I_{t}^{l}}|$ as input and produces a pixel-wise occlusion mask $O_{t}^{l}$ whose element denotes per-pixel occlusion probability from $0$ to $1$. Next, the reconstruction loss $L_{t}^{ar}$ is re-weighted by the occlusion mask $O_{t}^{l}$ and error map $e_{t}^{l}$ on each pixel, as shown in  Fig. \ref{fig-self_supervised}. Furthermore, we introduce the disparity smoothness loss ($L_{t}^{sm}$) to avoid possible artifacts. To guide the occlusion mask learning on the target domain more effectively, as shown in Fig. \ref{fig:training_diagram}, the weight-sharing occlusion prediction network simultaneously learns an occlusion mask $O_{s}^{l}$ on the source domain under the supervision of the ground-truth occlusion mask $\hat{O_{s}^{l}}$ generated from the ground-truth disparity map $\hat{d_{s}^{l}}$. Compared with the widely adopted  consistency check approaches (\emph{e.g.} left-right consistency check \cite{monodepth17}, per-pixel minimum reprojection \cite{godard2019digging}), our domain-collaborative occlusion mask learning scheme is more training-efficient, without the need of extra views of image and disparity map.
	
    As mentioned above, the occlusion prediction network simultaneously conducts occlusion mask learning on the source domain under the supervision of ground-truth occlusion mask $\hat{O_{s}^{l}}$, which however is not provided in the source-domain dataset. Here we specify how to obtain $\hat{O_{s}^{l}}$ given a source-domain ground-truth disparity map $\hat{d_{s}^{l}}$. Based on the binocular geometry, per-pixel occlusion judgment is defined as: 
    
    \begin{equation}
    \begin{aligned}
    \hat{O_{s}^{l}}(x,y)=\left\{
    \begin{array}{lr}
    \displaystyle 1 \quad \exists ~x_2>x,~ x_2-\hat{d_{s}^{l}}(x_2,y)=x-\hat{d_{s}^{l}}(x,y) \\
    \displaystyle 0 \quad else
    \end{array}
    \right.
    \end{aligned}
    \end{equation}
    where $\hat{O_{s}^{l}}(x,y)=1$ or $0$ denotes if the pixel $(x,y)$ is occluded or not, and $\hat{d_{s}^{l}}(x,y)$ denotes the source-domain ground-truth disparity at $(x,y)$.
    
    Our self-supervised occlusion-aware reconstruction task is the \emph{first proposed auxiliary task} for stereo domain adaptation. In addition, our method enables collaborative occlusion mask learning on both source and target domains, which acts like another domain adaptation sub-task on occlusion prediction that ensures the quality of target-domain occlusion masks, thereby explicitly improving the validity of reconstruction loss. Experimental results further validate the superiority of our proposed auxiliary task over other high-level auxiliary tasks.

   \subsection{Training Loss}
    
    On the source domain, we train the main task of disparity regression using the per-pixel smooth-\emph{L}1 loss (as same in \cite{chang2018pyramid}):  $L_{s}^{main} = Smooth_{L_{1}}(d_{s}^{l}-\hat{d_{s}^{l}})$. In addition, the per-pixel binary cross entropy loss is adopted for occlusion mask training on the source domain: $L_{s}^{occ} = BCE(O_{s}^{l},\hat{O_{s}^{l}})$.
    
    On the target domain, the occlusion-aware appearance reconstruction loss is defined as:
    \begin{equation}
    \begin{aligned}
    \, \, \, L_{t}^{ar} = \alpha\frac{1-SSIM(I_{t}^{l}\odot(1-O_{t}^{l}),\overline{I_{t}^{l}}\odot(1-O_{t}^{l}))}{2}  \\ 
    + \,\,\, (1-\alpha)||I_{t}^{l}\odot(1-O_{t}^{l})-\overline{I_{t}^{l}}\odot(1-O_{t}^{l})||_{1}
    \end{aligned}
    \end{equation}
    where $\odot$ denotes element-wise multiplication, $SSIM$ denotes a simplified single scale SSIM \cite{wang2004image} term with a $3\times3$ block fiter, and $\alpha$ is set to $0.85$. Besides, we apply a \emph{L}1-regularization term on the produced target-domain occlusion mask: $L_{t}^{occ} = ||O_{t}^{l}||_{1}$. Last but not least, we adopt an edge-aware term as the target-domain disparity smoothness loss, where $\partial{I}$ and $\partial{d}$ denote image and disparity gradients:
    \begin{equation}
    \\ \\ \, L_{t}^{sm} = |\partial_{x}{d_{t}^{l}}|e^{-|\partial_{x}{I_{t}^{l}}|} +  |\partial_{y}{d_{t}^{l}}|e^{-|\partial_{y}{I_{t}^{l}}|}
    \end{equation}
    
    Finally, the total training loss is a weighted sum of five loss terms above, where $\lambda_{s}^{occ}$, $\lambda_{t}^{ar}$, $\lambda_{t}^{occ}$ and $\lambda_{t}^{sm}$ denote corresponding loss weights:
    \begin{equation}
    \label{loss_t}
    \\ \, L = L_{s}^{main} + \lambda_{s}^{occ}L_{s}^{occ} + \lambda_{t}^{ar}L_{t}^{ar} + \lambda_{t}^{occ}L_{t}^{occ} + \lambda_{t}^{sm}L_{t}^{sm}
    \end{equation}

\section{Experiment}
    
    To prove the effectiveness of our method, we extend the $2$-D stereo baseline network ResNetCorr \cite{yang2018srcdisp} as \textbf{Ada-ResNetCorr} and the $3$-D stereo baseline network PSMNet \cite{chang2018pyramid} as \textbf{Ada-PSMNet}. We first conduct intensive ablation studies and detailed break-down comparisons for each proposed module on multiple datasets, including KITTI \cite{Geiger2012CVPR,Menze2015CVPR}, Middlebury \cite{Scharstein2014High}, ETH3D \cite{schoeps2017cvpr} and DrivingStereo \cite{yang2019drivingstereo}. Next, we compare the cross-domain performance of our method with other traditional / domain generalization / domain adaptation stereo methods. Then, we show that our domain-adaptive models achieve remarkable performance on four public stereo matching benchmarks, even outperforming several end-to-end disparity networks finetuned on target domains. Finally, we present additional verification experiments and applications of our proposed domain-adaptive stereo matching pipeline.

    \subsection{Datasets}
    
    
    \textbf{SceneFlow} dataset \cite{mayer2016large} is a large synthetic dataset containing $35k$ training pairs with dense ground-truth disparities, mainly acting as the source-domain training set. \textbf{KITTI} dataset includes two subsets: KITTI 2012 \cite{Geiger2012CVPR} and KITTI 2015 \cite{Menze2015CVPR}, each providing about $200$ stereo pairs of outdoor driving scenes with sparse ground-truth disparities for training and about $200$ image pairs for testing.  \textbf{Middlebury} dataset \cite{Scharstein2014High} is a small indoor dataset containing $15$ pairs for training and $15$ pairs for testing. Besides, we collect $57$ unused pairs from $5$ subsets (Middlebury 2001, 2003, 2005, 2006 and 2014), which are fused with all training pairs for unsupervised domain adaptation training. \textbf{ETH3D} dataset \cite{schoeps2017cvpr} includes both indoor and outdoor scenarios, containing $27$ gray image pairs with dense ground-truth disparities for training and $20$ image pairs for testing. \textbf{DrivingStereo} dataset \cite{yang2019drivingstereo} is a large-scale stereo matching dataset covering a diverse set of driving scenarios, containing $175K$ stereo pairs for training and $7751$ pairs for testing. These four real-world datasets mainly act as different target domains for domain adaptation evaluations.
    
    We adopt the bad pixel error ($D1$-error) as the evaluation metric, which calculates the percentage of pixels whose disparity errors are greater than a certain threshold ($1$-pixel error for ETH3D, $2$-pixel error for Middlebury, $3$-pixel error for KITTI and DrivingStereo).

    \subsection{Implementation Details}
    
    Each model is trained end-to-end using Adam optimizer ($\beta_1=0.9$, $\beta_2=0.999$) on eight NVIDIA Tesla-V100 GPUs. The learning rate is set to $0.001$ for training from scratch, and we train each model for $100$ epochs with the batch size of $16$ using $624\times304$ random crops. Referring to Tab. \ref{t9}, the momentum factor $\gamma$ in Alg. \ref{alg:color_transfer} is set to $0.95$. Referring to Tab. \ref{a1}, the weights of different loss terms ($\lambda_{s}^{occ}$, $\lambda_{t}^{ar}$, $\lambda_{t}^{occ}$,  $\lambda_{t}^{sm}$) in Eq.~\ref{loss_t} are set to ($0.2$, $1.0$, $0.2$, $0.1$). Considering some target-domain training sets are small, we apply several common data augmentation operations on target-domain training pairs only for the self-supervised occlusion-aware reconstruction (not for color transfer), including color shift, saturation, contrast adjustments and style-PCA based lighting noises.
    
    Next, we introduce the network structures of our domain-adaptive stereo models: Ada-ResNetCorr and Ada-PSMNet. Ada-ResNetCorr is extended based on the ResNetCorr \cite{yang2018srcdisp}, which is a baseline model among correlation-based $2$-D disparity networks. We extend the ResNetCorr, where ``conv1\_1" to ``conv1\_3'' in the ResNet-50 \cite{he2016deep} are adopted as the shallow feature extractor. We also replace the single-channel output convolutional layer with a soft-argmin block in $3$-D stereo networks \cite{kendall2017end,chang2018pyramid} for disparity regression. The proposed cost normalization layer is directly applied on the extracted lower-layer features of two views before correlation, which are of $1/2$ spatial size to the input image. The maximum displacement in the correlation layer is set to $128$.  Ada-PSMNet is extended based on the PSMNet \cite{chang2018pyramid}, which is a baseline model among cost-volume based $3$-D disparity networks. We follow the structure of PSMNet except the maximum disparity range is set to $256$. The cost normalization layer is directly applied on the extracted lower-layer features of two views before constructing the $4$-D cost volume.

    Then, we detail the structure of our occlusion prediction network (only adopted in training), which is a small fully-convolutional network. The first and second convolutional layers both use the $3\times3$ kernel (stride=$1$ with $32$ output channels), followed by a batch normalization layer and a ReLU layer each. The last convolutional layer uses the $1\times1$ kernel (stride=$1$ with a single output channel), followed by a sigmoid layer, producing a full-size occlusion mask whose element denotes per-pixel occlusion probability from $0$ to $1$.

    Finally, we reiterate our training and testing protocols. An individual domain-adaptive stereo model is trained for each target domain, and we do not use any images from the target-domain test set during training. For fair comparisons, all experiments are conducted rigorously following the same unsupervised domain adaptation settings: training on the source domain also with target-domain images (no target-domain labels), then directly testing on the target domain. Due to the number of submissions to online test servers is strictly limited, we conduct ablation studies and tune hyperparameters on various target-domain training sets.
    
    \begin{figure*}[tb]
    	\centering
    	\includegraphics[width=0.85\textwidth]{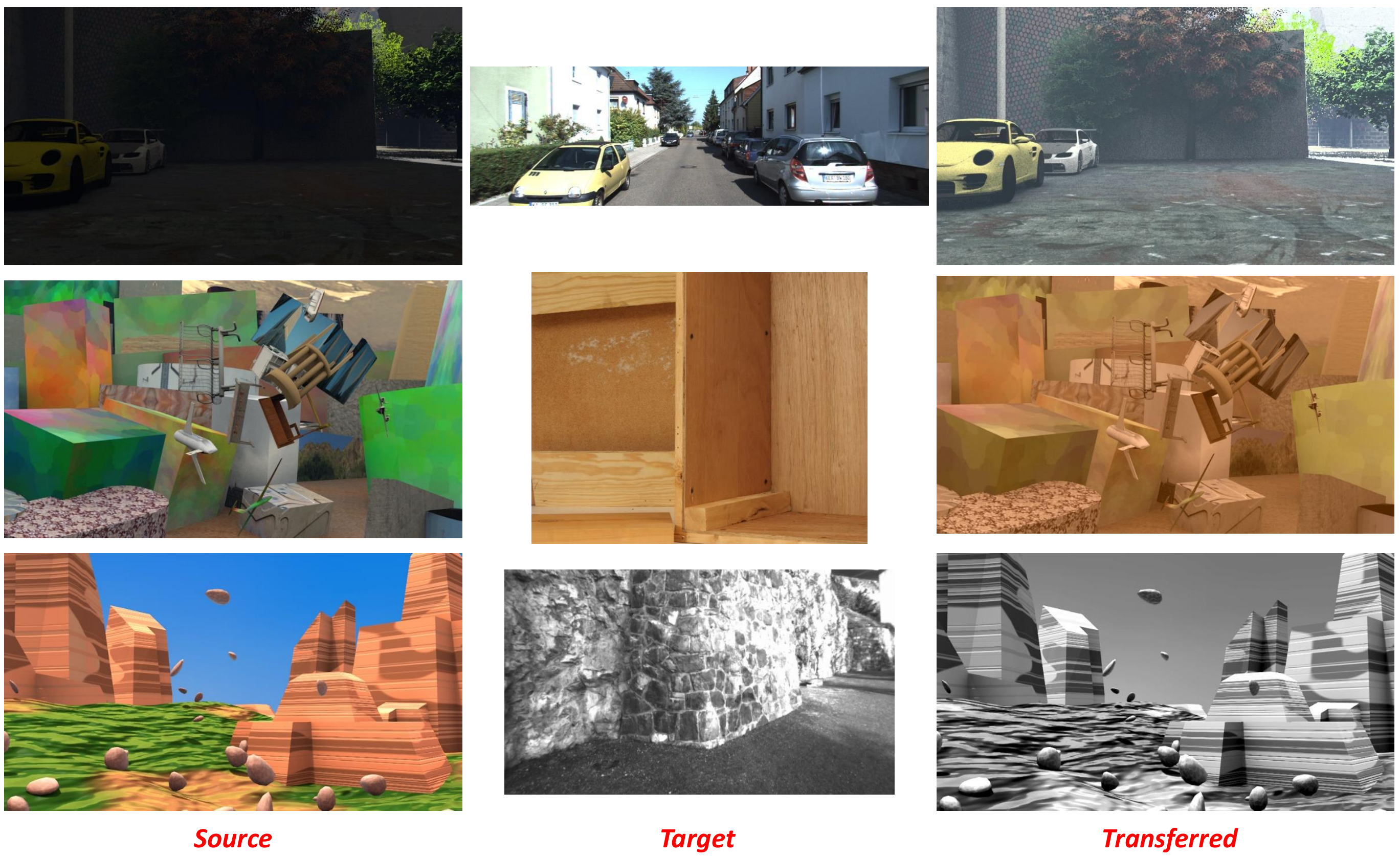}
    	\caption{
    		Qualitative results of color transfer 	from synthetic (SceneFlow) to real-world scenes. Top-down: transfer to KITTI, Middlebury and ETH3D. As can be seen, our non-adversarial color transfer method enables highly effective image style translations, and ensures semantic invariance without any geometrical distortion, which is of vital importance for the low-level stereo matching task.
	    }
    	\label{fig:color_trans}
    \end{figure*}
    
    \begin{table*}[htb]
    	\centering
    	\caption{Ablation studies on the KITTI, Middleburry, ETH3D and DrivingStereo training sets. $D1$-error (\%) is adopted for evaluations. All models are trained on the synthetic SceneFlow dataset.}
    	\vspace{-2.5mm}
    	\resizebox{0.99\textwidth}{!}{
    		\begin{tabular}{c | c c c | c c | c c | c | c }
    			\hline
    			\multirow{2}{*}{Model} & {cost} & {color} & {self-supervised} & 
    			\multicolumn{2}{c|}{~~~~~~~~KITTI~~~~~~~} & 
    			\multicolumn{2}{c|}{~~~~Middlebury~~~~} & 
    			\multirow{2}{*}{~~~ETH3D~~~} & 
    			\multirow{2}{*}{~DrivingStereo~} \\
    			{} & {normalization} & {transfer} & {reconstruction} & 
    			{2012} & {2015} & {half} & {quarter} & {} & \\
    			\hline
    			\multirow{5}{*}{\adapsmnet} & 
    			{\xmark} & {\xmark} & {\xmark} & 
    			13.6 & 12.1 & 18.6 & 11.5 & 10.8 & 20.9 \\
    			{} & {\cmark} & {\xmark} & {\xmark} & 
    			11.8 & 9.1 & 16.8 & 10.1 & 9.0 & 16.7 \\
    			{} & {\xmark} & {\cmark} & {\xmark} & 
    			5.3 & 5.4 & 10.0 & 5.8 & 6.1 & 7.4 \\
    			{} & {\cmark} & {\cmark} & {\xmark} & 
    			4.5 & 4.7 & 9.0 & 5.1 & 5.2 & 6.4 \\
    			{} & {\cmark} & {\cmark}  & {\cmark} & 
    			\textbf{3.6} & \textbf{3.5} & \textbf{8.4} & \textbf{4.7} & \textbf{4.1} & \textbf{5.1} \\
    			\hline
    			\multirow{5}{*}{\adaresnetcorr} & 
    			{\xmark} & {\xmark} & {\xmark} & 
    			9.8 & 9.4 & 22.5 & 12.8 & 15.8 & 17.2 \\
    			{} & {\cmark} & {\xmark} & {\xmark} & 
    			8.1 & 8.4 & 19.7 & 10.9 & 13.4 & 15.2 \\
    			{} & {\xmark} & {\cmark} & {\xmark} & 
    			6.7 & 6.7 & 15.1 & 8.3 & 7.1 & 10.2 \\
    			{} & {\cmark} & {\cmark} & {\xmark} & 
    			6.0 & 5.9 & 13.7 & 7.4 & 6.6 & 9.2 \\
    			& {\cmark} & {\cmark} & {\cmark} & 
    			\textbf{5.1} & \textbf{5.0} & \textbf{12.7} & \textbf{6.6} & \textbf{5.8} & \textbf{8.0} \\
    			\hline
    		\end{tabular}
    	}
    	\vspace{-5pt}
    	\label{t1}
    \end{table*}
    
    \subsection{Ablation Studies}
    \label{subsec:ablation}
    
    \subsubsection{Basic Ablation Studies}
    
    In Tab. \ref{t1}, we conduct detailed ablation studies on four real-world datasets to evaluate the key components in our domain adaptation pipeline, based on Ada-PSMNet and Ada-ResNetCorr. As can be seen, applying our non-adversarial progressive color transfer method in training can significantly reduce error rates on multiple target domains, \emph{e.g.} $8.3\%$ on KITTI, $8.6\%$ on Middlebury, $4.7\%$ on ETH3D and $13.5\%$ on DrivingStereo for Ada-PSMNet, benefiting from massive color-aligned training images without geometrical distortions. We also provide qualitative results of our color transfer method in Fig. \ref{fig:color_trans}. In addition, compared with baseline models, the error rates are reduced by $1\%\sim4\%$ on target domains by integrating the proposed cost normalization layer, which also works well when implemented together with the input color transfer module. Furthermore, adopting the self-supervised occlusion-aware reconstruction can further reduce error rates by $0.5\%\sim1.5\%$ on various target domains, despite the adaptation performance is already remarkable after color transfer and cost normalization. Finally, both Ada-PSMNet and Ada-ResNetCorr outperform the corresponding baseline model by notable margins on all target domains, especially an accuracy improvement of $15.8\%$ by Ada-PSMNet on the large-scale DrivingStereo training set.
    
    
    \begin{table}[h]
    	\centering
    	\caption{Sensitivity tests of the momentum factor $\gamma$ in our progressive color transfer algorithm on the KITTI and DrivingStereo training sets. $D1$-error (\%) is adopted for evaluations.}
    	\vspace{-2.5mm}
    	\resizebox{0.92\linewidth}{!}{
    		\begin{tabular}{ c | c  c  c  c c c  }
    			\hline
    			Target Domain  & {$\gamma$=1} & {0.975} & {0.95} & {0.9} & {0.8} & {0.0} \\
    			\hline
    			KITTI  & 6.2 & 5.8 & 5.4 & 5.6 & 5.9 & 9.3 \\
    			DrivingStereo & 8.3 & 7.8 & 7.4 & 7.6 & 8.0 & 12.4  \\
    			\hline
    		\end{tabular}
    	}
    	\label{t9}
    \end{table}

     \begin{table}[h]
     	\centering
     	\caption{Supplementary ablation studies for cost normalization on the KITTI and DrivingStereo training sets. $D1$-error (\%) is adopted for evaluations.}
     	\vspace{-2.5mm}
     	\resizebox{1\linewidth}{!}{
     		\begin{tabular}{ c | c  c  c  c  }
     			\hline
     			Target Domain  & {Baseline} & {+Channel Norm} & {+Pixel Norm} & {+Cost Norm} \\
     			\hline
     			KITTI  & 12.1 & 11.0 & 10.5 & 9.1 \\
     			DrivingStereo & 20.9 & 19.4 & 18.7 & 16.7 \\
     			\hline
     		\end{tabular}
     	}
     	\label{t10}
     \end{table}

    \begin{table}[h]
    	\centering
    	\caption{Supplementary ablation studies for different loss terms in self-supervised occlusion-aware reconstruction on the KITTI and DrivingStereo training sets. $D1$-error (\%) is adopted for evaluations.}
    	\vspace{-2.5mm}
    	\resizebox{1\linewidth}{!}{
    		\begin{tabular}{ c c c c c | c c }
    			\hline
    			{$L_{s}^{main}$} & {$L_{s}^{occ}$} & {$L_{t}^{ar}$} & {$L_{t}^{occ}$} & {$L_{t}^{sm}$} & {KITTI} & {DrivingStereo} \\
    			\hline
    			{\cmark} & {\xmark} & {\xmark} & {\xmark} & {\xmark} & 4.7 & 6.4 \\
    			{\cmark} & {\xmark} & {\cmark} & {\xmark} & {\xmark} & 3.9 & 5.6 \\
    			{\cmark} & {\cmark} & {\cmark} & {\cmark} & {\xmark} & 3.6 & 5.2 \\
    			{\cmark} & {\cmark} & {\cmark} & {\cmark} & {\cmark} & 3.5 & 5.1 \\
    			{\cmark} & {\xmark} & {\cmark} & {\cmark} & {\cmark} & 3.8 & 5.4 \\
    			\hline
    		\end{tabular}
    	}
    	\label{t11}
    \end{table}
    
    For further demonstrations of effectiveness inside each proposed module, we provide several detailed ablation studies. In Tab. \ref{t9}, we conduct the sensitivity test of the momentum factor $\gamma$ in our progressive color transfer algorithm. According to Line 6-7 in Alg. \ref{alg:color_transfer}, $\gamma$=$1$ denotes no progressive update scheme in color transfer, while the error rates on target domains begin to fall with the decrease of $\gamma$. When $\gamma$ is set to $0.95$, the best performance on target domains is achieved, indicating that the progressive update scheme simultaneously ensures the diversity and representativeness of target-domain color styles during style transfer, thereby benefiting the stereo domain adaptation. Besides, our progressive color transfer algorithm shows good robustness to the momentum factor $\gamma$. In Tab. \ref{t10}, we conduct detailed ablation studies for two proposed normalization operations in the cost normalization layer. Applying the channel normalization or pixel normalization alone, can reduce the error rate on the KITTI training set by $1.1\%$ or $1.6\%$ respectively, indicating that the pixel normalization is more important to some degree. When combining together, the performance gain is greater than the sum of two individual gains, demonstrating that these two operations are complementary. In Tab. \ref{t11}, we conduct detailed ablation studies for different loss terms in self-supervised occlusion-aware reconstruction. When progressively applying the reconstruction loss ($L_{t}^{ar}$), occlusion regularization loss ($L_{s}^{occ}$ \& $L_{t}^{occ}$) and disparity smoothness loss ($L_{t}^{sm}$), the error rates on the target-domain DrivingStereo training set are reduced by $0.8\%$, $0.4\%$ and $0.1\%$. Besides, the domain-collaborative occlusion mask learning is also important, manifesting as an increase of $0.3\%$ in $D1$-error if the source-domain occlusion mask training loss $L_{s}^{occ}$ is not adopted. In Tab. \ref{a1}, we tune different combinations of training loss weights on the KITTI training set. As can be seen, our domain-adaptive training pipeline is robust to these hyperparameters.

    \begin{table}[t]
    \centering
        \caption{Ablation studies for training loss weights on the KITTI training set. $D1$-error (\%) is adopted for evaluations.}
        \vspace{-2.5mm}
        \resizebox{0.8\linewidth}{!}{
            \begin{tabular}{ c | c c c c | c }
            \hline
           {$L_{s}^{main}$} & {$\lambda_{t}^{ar}$} & {$\lambda_{s}^{occ}$} &  {$\lambda_{t}^{occ}$} & {$\lambda_{t}^{sm}$}  & {KITTI} \\
          \hline
          {\cmark} & 0.0 & 0.0 & 0.0 & 0.0 & 4.7  \\
          {\cmark} & 0.8 & 0.0 & 0.0 & 0.0 & 4.0  \\
          {\cmark} & 1.0 & 0.0 & 0.0 & 0.0 & 3.9  \\
          {\cmark} & 1.2 & 0.0 & 0.0 & 0.0 & 4.1  \\
          {\cmark} & 1.0 & 0.1 & 0.1 & 0.0 & 3.7  \\
          {\cmark} & 1.0 & 0.2 & 0.2 & 0.0 & 3.6  \\
          {\cmark} & 1.0 & 0.3 & 0.3 & 0.0 & 3.7  \\
          {\cmark} & \textbf{1.0} & \textbf{0.2} & \textbf{0.2} & \textbf{0.1} & \textbf{3.5}  \\
          {\cmark} & 1.0 & 0.2 & 0.2 & 0.2 & 3.6  \\
            \hline
            \end{tabular}
        }
        \vspace{-8pt}
    \label{a1}
    \end{table}

	\begin{table}[b]
		\centering
		\caption{
			Comparisons with existing normalization layers on the KITTI and DrivingStereo training sets. $D1$-error (\%) is adopted for evaluations. 
		}
		\vspace{-2.5mm}
		\resizebox{0.9\linewidth}{!}{
			\begin{tabular}{c | c  c}
				\hline
				{Methods} & 
				{KITTI} & 
				{DrivingStereo} \\
				\hline
				{PSMNet Baseline}    & 12.1 & 20.9 \\
				{+Adaptive Batch Norm \cite{li2016revisiting}}    & 11.8 & 20.3 \\
				{+Batch-Instance Norm \cite{nam2018batch}}    & 11.2 & 19.5 \\
				{+Instance Norm \cite{ulyanov2016instance}} & 10.7 & 18.6 \\
				{+Domain Norm \cite{zhang2020domain}} & 9.5 & 17.2 \\
				{+Our Cost Norm} & 9.1 & 16.7 \\
				\hline
			\end{tabular}
		}
		\label{t6}
	\end{table}
	
    \subsubsection{Break-down Comparisons of Each Module}
    
    In order to further demonstrate the superiority of each module for domain-adaptive stereo matching, we perform exhaustive comparisons with other alternative methods respectively. As shown in Tab.~\ref{t6}, our specifically designed cost normalization layer which is parameter-free and adopted only once in the network, outperforms other general and learnable normalization layers (AdaBN \cite{li2016revisiting}, BIN \cite{nam2018batch}, IN \cite{ulyanov2016instance} and DN \cite{zhang2020domain}) which are repeatedly adopted in the network's feature extractor. In Tab.  \ref{t7}, our progressive color transfer algorithm far outperforms four popular color/style transfer networks  (WCT$^{2}$~\cite{yoo2019photorealistic}, WaterGAN \cite{li2017watergan}, CycleGAN \cite{zhu2017unpaired} and StereoGAN \cite{liu2020stereogan}), indicating that geometrical distortions from such GAN-based color/style transfer models are harmful for the low-level stereo matching task. Moreover, our method outperforms the Reinhard's color transfer method \cite{reinhard2001color} by about $1\%$ in $D1$-error, revealing the effectiveness of the proposed progressive update scheme. In Tab.  \ref{t8}, our proposed self-supervised occlusion-aware reconstruction task is demonstrated to be a highly effective auxiliary task for domain-adaptive stereo matching, while other alternatives all hurt the domain adaptation performance.

    \begin{table}[t]
    	\centering
    	\caption{
    		Comparisons with other color/style transfer methods on the KITTI and DrivingStereo training sets. $D1$-error (\%) is adopted for evaluations. 
    	}
    	\vspace{-2.5mm}
    	\resizebox{0.99\linewidth}{!}{
    		\begin{tabular}{c | c  c}
    			\hline
    			{Methods} & 
    			{KITTI} & 
    			{DrivingStereo} \\
    			\hline
    			{PSMNet Baseline}    & 12.1 & 20.9 \\
    			{+WCT$^{2}$ \cite{yoo2019photorealistic}} & 10.2 & 17.3 \\
    			{+WaterGAN \cite{li2017watergan}} & 8.7 & 11.5 \\
    			{+CycleGAN \cite{zhu2017unpaired}} & 8.0 & 10.6 \\
    			{+StereoGAN \cite{liu2020stereogan}} & 6.6 & 8.8 \\
    			{+Color Transfer \cite{reinhard2001color}} & 6.2 & 8.3 \\
    			{+Our Progressive Color Transfer} & 5.4 & 7.4 \\
    			\hline
    		\end{tabular}
    	}
    	\label{t7}
    \end{table}

    \begin{table}[t]
    	\centering
    	\caption{
    		Comparisons with other auxiliary tasks for stereo domain adaptation on the KITTI and DrivingStereo training sets. $D1$-error (\%) is adopted for evaluations.
    	}
    	\vspace{-2.5mm}
    	\resizebox{1\linewidth}{!}{
    		\begin{tabular}{c | c  c}
    			\hline
    			Methods & KITTI & DrivingStereo  \\
    			\hline
    			PSMNet Baseline    & 12.1 & 20.9 \\
    			+Patch Localization Task \cite{xu2019self} & 14.2 & 23.5 \\
    			+Rotation Prediction Task \cite{gidaris2018unsupervised} & 14.6 & 24.6  \\
    			+Flip Prediction Task \cite{xu2019self} & 15.1 & 24.3  \\
    			+Our Self-Supervised Reconstruction Task & 7.6 & 12.3 \\
    			\hline
    			PSMNet + Our Color Transfer + Cost Norm & 4.7 & 6.4 \\
    			+Patch Localization Task \cite{xu2019self} & 6.5 & 8.0 \\
    			+Rotation Prediction Task \cite{gidaris2018unsupervised} & 6.1 & 8.4 \\
    			+Flip Prediction Task \cite{xu2019self} & 6.0 & 8.3  \\
    			+Our Self-Supervised Reconstruction Task & 3.5 & 5.1 \\
    			\hline
    		\end{tabular}
    	}
    	\label{t8}
    \end{table}
    
    \begin{table}[t]
    	\centering
    	\caption{
    		Ablation studies for different amounts of target-domain stereo pairs used in training, on the KITTI, Middlebury (half) and DrivingStereo training sets. Ada-PSMNet is adopted for comparisons.
    	}
    	\vspace{-2.5mm}
    	\resizebox{1\linewidth}{!}{
    		\begin{tabular}{c | c c  c c c}
    			\hline
    			\multicolumn{6}{c}{DrivingStereo} \\
    			\hline
    			Sampling ratio & 0 & 1/5 & 1/2 & 1  \\
    			Target-domain training pairs & 0 & 35000 & 87500 & 175000 \\
    			$D1$-error (\%) & 20.9 & 5.4 & 5.2 & 5.1 \\
    			\hline
    			\multicolumn{6}{c}{KITTI} \\
    			\hline
    			Sampling ratio & 0 & 1/2  & 3/4  & 1  \\
    			Target-domain training pairs & 0 & 100 & 150 & 200 \\
    			$D1$-error (\%) & 12.1 & 4.6 & 3.8 & 3.5 \\
    			\hline
    			\multicolumn{6}{c}{Middlebury} \\
    			\hline
    			Sampling ratio & 0  & 1/2  & 3/4  & 1  \\
    			Target-domain training pairs & 0 & 36 & 54 & 72 \\
    			$D1$-error (\%) & 18.6 & 9.9 & 9.0 & 8.4 \\
    			\hline
    		\end{tabular}
    	}
    	\vspace{-10pt}
    	\label{t13}
    \end{table}

    \subsubsection{Robustness to the Amount of Target-domain Training Samples}
    
    In Tab. \ref{t13}, we conduct detailed ablation studies for different amounts of target-domain images used in training, to validate the robustness of our domain adaptation pipeline to the amount of target-domain training samples. The experiments are conducted on three target-domain training sets with different scales: DrivingStereo ($175k$ pairs), KITTI ($200$ pairs) and Middlebury ($72$ pairs). For the large-scale DrivingStereo dataset, even if we only use one fifth of the training samples for unsupervised domain adaptation training, a remarkable $D1$-error of $5.4\%$ is achieved on the whole training set, which is only inferior to that of using all samples for training by $0.3\%$. The performance gap is further narrowed to $0.1\%$ when a half of the training samples are used. For small target-domain datasets such as KITTI and Middlebury, using only half of the training samples can reduce error rates on corresponding training sets by $7.5\%$ and $8.7\%$ respectively. When the sampling ratio increases to $3/4$, the performance gaps to using all samples for training are narrowed to $0.3\%$ and $0.6\%$ respectively. To sum up, our proposed domain adaptation pipeline is robust to varied capacities of target-domain datasets.

   \begin{table}[t]
    \centering
    \caption{
        Cross-domain comparisons with other traditional / domain generalization / domain adaptation stereo methods on the KITTI / Middlebury / ETH3D training sets, and DrivingStereo test set. $D1$-error (\%) is adopted for evaluations. The second and third columns indicate whether the method is trained on the SceneFlow dataset, and whether the method uses target-domain images during training respectively. 
    }
    \vspace{-2.5mm}
    \resizebox{1\linewidth}{!}{
        \begin{tabular}{c | c | c | c  c  c  c }
	        \hline
            \multirow{2}{*}{Methods} & 
            {Train} & 
            {Target} &
            \multicolumn{4}{c}{Test} \\
            {} & 
            {SceneFlow} & 
            {Images} &
            {KIT} & 
            {Mid} & 
            {ETH} &
            {DS}\\
            \hline
            \multicolumn{7}{c}{Traditional Stereo Methods} \\
            \hline
            {PatchMatch \cite{bleyer2011patchmatch}}    & {\xmark} & {\xmark} & 17.2 & 38.6 & 24.1 & - \\
            {SGM \cite{hirschmuller2008stereo}}           & {\xmark} & {\xmark} & 7.6 & 25.2  & 12.9 & 26.4  \\
            \hline
            \multicolumn{7}{c}{Domain Generalization Stereo Networks} \\
            \hline
            {HD$^{3}$ \cite{yin2019hierarchical}}      & {\cmark} & {\xmark} & 26.5  & 37.9  & 54.2 & 33.1 \\
            {GWCNet \cite{guo2019group}}        & {\cmark} & {\xmark} & 22.7  & 34.2  & 30.1 & 26.9 \\
            {PSMNet \cite{chang2018pyramid}}        & {\cmark} & {\xmark} & 16.3  & 25.1  & 23.8 & 21.4 \\
            {GANet \cite{zhang2019ga}}         & {\cmark} & {\xmark} & 11.7  & 20.3  & 14.1 & 16.8 \\
            {DSMNet \cite{zhang2020domain}}        & {\cmark} & {\xmark} & 6.5   & 13.8  & 6.2 & - \\
            \hline
            \multicolumn{7}{c}{Domain Adaptation Stereo Networks} \\
            \hline
            {StereoGAN \cite{liu2020stereogan}}      & {\cmark} & {\cmark} & 12.1  & -     & - & - \\
            {\adaresnetcorr} & {\cmark} & {\cmark} & 5.0 & 12.7 & 5.8 & 7.8 \\
            {\adapsmnet}     & {\cmark} & {\cmark} & \textbf{3.5} & \textbf{8.4} & \textbf{4.1} & \textbf{5.2} \\
            \hline
        \end{tabular}
    }
    \label{t2}
    \end{table}

    \subsection{Cross-domain Comparisons}
    
    
    \begin{figure*}[t]
        \centering
        \includegraphics[width=0.85\linewidth]{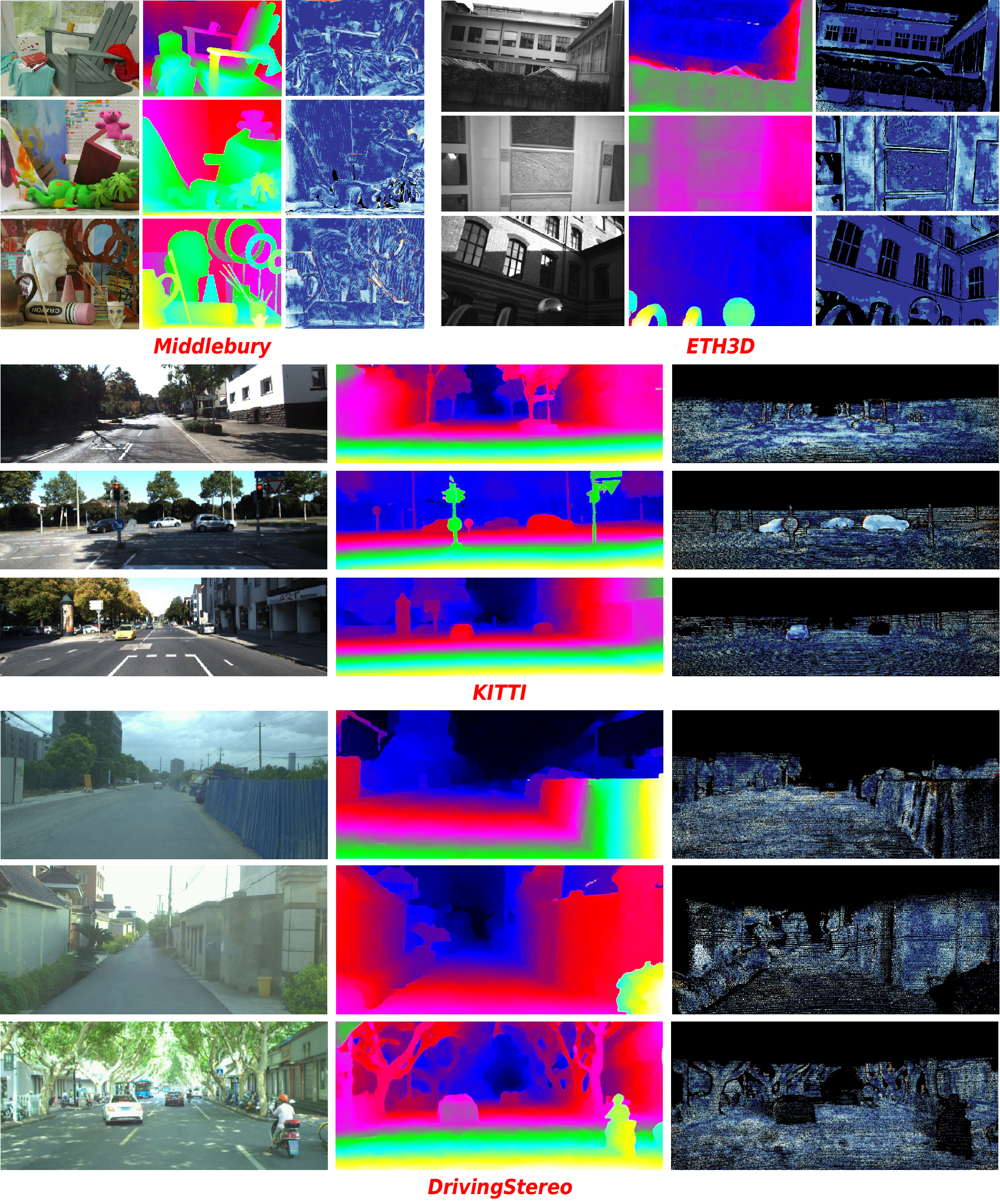}
        	\caption{Disparity predictions from our SceneFlow-pretrained Ada-PSMNet on the KITTI, ETH3D, Middlebury and DrivingStereo datasets. Left-right: left stereo image, colorized disparity and error maps.
        }
        \vspace{-8pt}
        \label{fig:disp_pred} 
    \end{figure*}
    
    In Tab. \ref{t2}, we compare our proposed domain-adaptive stereo models with other traditional stereo methods, domain generalization, and domain adaptation stereo networks on four real-world datasets for cross-domain comparisons. 
    
    \begin{itemize}
    \item[\small{$\bullet$}] Both Ada-ResNetCorr and Ada-PSMNet show great superiority over traditional  methods on all target domains.
    
    \item[\small{$\bullet$}] For comparisons with domain generalization networks, unfairness may exist since our domain-adaptive models use target-domain images during training. It is caused by the problem definition of \textbf{Domain Adaptation} as mentioned in Sec. \ref{sec:formulation}. However, as shown in Tab.~\ref{t2}, our Ada-PSMNet achieves tremendous gains rather than small deltas compared with all domain generalization stereo networks, including the state-of-the-art DSMNet \cite{zhang2020domain} and its baseline network GANet \cite{zhang2019ga}. For more comparisons with DSMNet~\cite{zhang2020domain}, Zhang \emph{et al.} reported an error rate of $8.5\%$ on the KITTI 2015 training set by applying their proposed domain normalization layer and non-local filter layers on the PSMNet, while our Ada-PSMNet achieves an error rate of $3.5\%$ using the same baseline network.
    
    \item[\small{$\bullet$}] For more ``fairer" comparisons with domain generalization networks on unseen target-domain scenes, we conduct further experiments on the DrivingStereo test set. As shown in Tab.~\ref{t2}, our domain-adaptive stereo models show significant superiority over other SceneFlow-pretrained models, as well as the classical method SGM.
    
    \item[\small{$\bullet$}] Among few published domain adaptation networks, only  StereoGAN \cite{liu2020stereogan} reported such cross-domain performance, while  Ada-PSMNet achieves a $3.5$ times lower error rate than StereoGAN~\cite{liu2020stereogan} on the KITTI training set.
    
    \end{itemize}

    Hence, our proposed multi-level alignment pipeline successfully address the domain adaptation problem for stereo matching. In Fig. \ref{fig:disp_pred}, we provide qualitative results of our method on different real-world datasets, in which accurate disparity maps are predicted on outdoor and indoor scenes.

    \begin{table*}[!t]
    \centering
        \caption{Performance on the ETH3D online test set. $1$-pixel and $2$-pixel errors (\%) are adopted for evaluations.}
        \vspace{-2.5mm}
        \resizebox{0.92\textwidth}{!}{
            \begin{tabular}{ c | c  c  c  c  c  c  c }
            \hline
            {Method} & {Deep-Pruner} & {iResNet} & {SGM-Forest} & {PSMNet} & {Stereo-DRNet} & {DispNet} & \textbf{{\adapsmnet}}\\
            {Use ETH-gt} & {\cmark} & {\cmark} & {\xmark} & {\cmark} & {\cmark} & {\cmark} & {\xmark} \\
            \hline
            {Bad 1.0} & 3.52 & 3.68 & 4.96 & 5.02 & 5.59 & 17.47 & \textbf{3.09} \\
            {Bad 2.0} & 0.86 & 1.00 & 1.84 & 1.09 & 1.48 & 7.91 & \textbf{0.65} \\
            \hline
            \end{tabular}
        }
    \label{t3}
    \end{table*}
    
    \begin{table*}[!t]
    \centering
        \caption{Performance on the Middlebury online test set. $2$-pixel error (\%) is adopted for evaluations.}
        \vspace{-2.5mm}
        \resizebox{0.92\textwidth}{!}{
            \begin{tabular}{ c | c  c  c  c  c  c  c }
            \hline
            {Method} & {EdgeStereo} & {CasStereo} & {iResNet} & {MCV-MFC} & {Deep-Pruner} & {PSMNet} & \textbf{\adapsmnet}\\
            {Use Mid-gt} & {\cmark} & {\cmark} & {\cmark} & {\cmark} & {\cmark} & {\cmark} & {\xmark} \\
            \hline
            Bad 2.0 & 18.7 & 18.8 & 22.9 & 24.8 & 30.1 & 42.1 & \textbf{13.7} \\
            \hline
            \end{tabular}
        }
    \label{t4}
    \end{table*}
	
	\begin{table*}[!t]
    \centering
        \caption{Performance on the KITTI 2015 online test set. $3$-pixel error (\%) is adopted for evaluations.}
        \vspace{-2.5mm}
        \resizebox{0.99\textwidth}{!}{
            \begin{tabular}{ c | c  c  c  c  c  c  c  c  c}
            \hline
            {Method} & {GC-Net} & {L-ResMatch} & {SGM-Net} & {MC-CNN} & {DispNetC}  & {Weak-Sup} & {MADNet} & {Unsupervised} & \textbf{{\adapsmnet}}\\
            {Use KITTI-gt} & {\cmark} & {\cmark} & {\cmark} & {\cmark} & {\cmark} & {\cmark} & {\xmark} & {\xmark} & {\xmark} \\
            \hline
            $D1$-error & \textbf{2.87} & 3.42 & 3.66 & 3.89 & 4.34 & 4.97 & 8.23 & 9.91 & 3.08 \\
            \hline
            \end{tabular}
        }
    \label{t5}
    \end{table*}
    
    \begin{table*}[!t]
    \centering
        \caption{Performance on the DrivingStereo test set. $3$-pixel error (\%) is adopted for evaluations.}
        \vspace{-2.5mm}
        \resizebox{0.90\textwidth}{!}{
            \begin{tabular}{ c | c  c  c  c c c  c  c }
            \hline
            {Method} & {iResNet} & {SegStereo} & {CRL} & {ResNetCorr} & {GuideNet} & {DispNetC} & {SGM} & \textbf{\adapsmnet}\\
            {Use Driving-gt} & {\cmark} & {\cmark} & {\cmark} & {\cmark} & {\cmark} & {\cmark} & {\xmark} & {\xmark} \\
            \hline
            $D1$-error & \textbf{4.27} & 5.89 & 6.02 & 6.75 & 8.89 & 16.82 & 26.44 & 5.22 \\
            \hline
            \end{tabular}
        }
    \label{t12}
    \end{table*}

    \subsection{Evaluations on Public Stereo Benchmarks}

    To further demonstrate the effectiveness of our method, we compare our domain-adaptive model Ada-PSMNet with several unsupervised/self-supervised methods and finetuned disparity networks on public stereo matching benchmarks: KITTI, Middlebury, ETH3D and DrivingStereo. \emph{\textbf{We directly upload the results from our SceneFlow-pretrained model to the online benchmark and do not finetune using target-domain ground-truths before submitting test results}}.

	\subsubsection{Results on the ETH3D Benchmark}
	
    As can be seen in Tab.  \ref{t3}, our proposed Ada-PSMNet outperforms the state-of-the-art patch-based network DeepPruner \cite{duggal2019deeppruner} and end-to-end disparity networks (iResNet \cite{liang2017learning}, PSMNet \cite{chang2018pyramid} and StereoDRNet \cite{chabra2019stereodrnet}) finetuned with ground-truth disparities in the ETH3D training set, as well as the SOTA traditional method SGM-Forest \cite{schonberger2018learning}. By the time of the paper submission, AdaStereo ranks 1$^{st}$ on the ETH3D benchmark in all evaluation metrics among published stereo matching methods.

    \subsubsection{Results on the Middlebury Benchmark}
    
    As can be seen in Tab.~\ref{t4}, the synthetic-data pretrained Ada-PSMNet outperforms all other state-of-the-art end-to-end disparity networks (EdgeStereo \cite{song2020edgestereo}, CasStereo \cite{gu2020cascade}, iResNet \cite{liang2017learning}, MCV-MFC \cite{liang2019stereo} and PSMNet \cite{chang2018pyramid}) which are finetuned using ground-truth disparities in the Middlebury training set by a noteworthy margin. Our Ada-PSMNet achieves a remarkable $2$-pixel error rate of $13.7\%$ on the full-resolution test set, outperforming all other finetuned end-to-end stereo matching networks on the benchmark.
    
    \subsubsection{Results on the KITTI Benchmark}
    
    As can be seen in Tab.  \ref{t5}, our domain-adaptive model far outperforms the online-adaptive model MADNet \cite{tonioni2019real}, the weak-supervised \cite{tulyakov2017weakly} and unsupervised \cite{zhou2017unsupervised} methods, meanwhile achieving higher accuracy than some supervised disparity networks including  MC-CNN-acrt \cite{zbontar2015computing},  L-ResMatch \cite{shaked2016improved}, DispNetC \cite{mayer2016large} and SGM-Net \cite{seki2017sgm}. Moreover, the SceneFlow-pretrained Ada-PSMNet achieves comparable performance with the KITTI-finetuned GC-Net \cite{kendall2017end}.
    
    \subsubsection{Results on the DrivingStereo Test Set}
    
    As can be seen in Tab. \ref{t12}, the SceneFlow-pretrained Ada-PSMNet far outperforms the traditional stereo method SGM \cite{hirschmuller2008stereo}, meanwhile achieving higher accuracy than some end-to-end disparity networks including SegStereo \cite{yang2018SegStereo}, CRL \cite{pang2017cascade}, ResNetCorr \cite{yang2018srcdisp}, GuideNet \cite{yang2019drivingstereo} and DispNetC \cite{mayer2016large}, which are all trained on the large-scale DrivingStereo training set with ground-truth disparity maps. Moreover, the SceneFlow-pretrained Ada-PSMNet achieves comparable performance with the DrivingStereo-trained iResNet \cite{liang2017learning}.
    
    \begin{figure*}[t]
    	\centering
    	\includegraphics[width=1\linewidth]{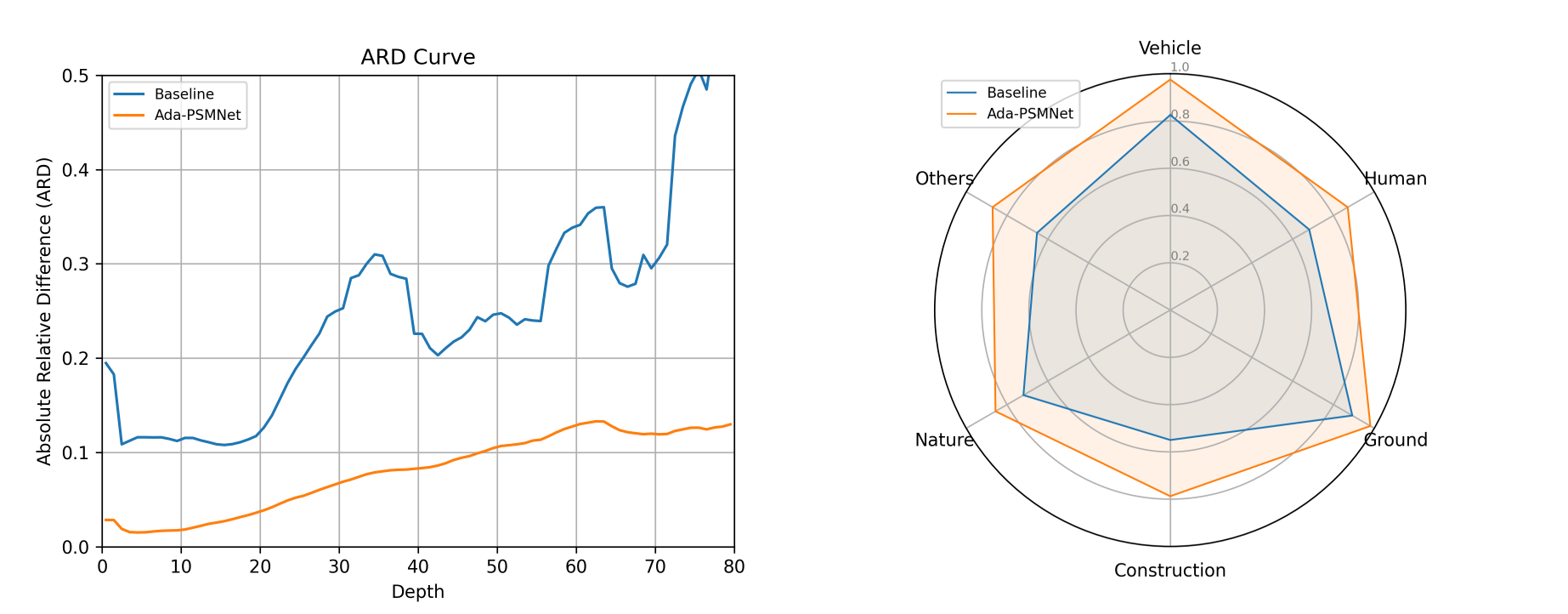}
    	\caption{Distance-aware ARD curve and semantic-aware radar map of the SceneFlow-pretrained PSMNet and Ada-PSMNet on the KITTI 2015 training set.
    	}
    	\label{fig-ard_curve}
    \end{figure*}
    
    \subsection{More Verification and Novel Applications}
    \label{subsec:apply}
    
    \subsubsection{Semi-supervised Applications}
    
    We conduct additional ``semi-supervised" experiments, to further validate the effectiveness of our domain adaptation method, when a small faction of target-domain ground-truth disparity maps are available for finetuning. Under this circumstance, target-domain test sets are more suitable for evaluations. Considering the DrivingStereo dataset is the only public stereo matching dataset with available test set labels, we choose DrivingStereo as the target domain in this part of ``semi-supervised" experiments. Specifically, we randomly select $500$ labeled pairs from the DrivingStereo training set (about $0.3\%$ of the total training set) to finetune our domain-adaptive models, then report results on the DrivingStereo test set. As can be seen in Tab. \ref{a2}, finetuning our domain-adaptive models with only $500$ target-domain labeled pairs, achieves the performance on par with training from scratch on the whole target-domain training set ($175K$ pairs) for $100$ epochs, which is also significantly better than finetuning the SceneFlow-pretrained models with the same amount of target-domain labeled pairs.

\begin{table}[t]
    \centering
        \caption{``Semi-supervised" performance on the DrivingStereo test set. ``SF" denotes the SceneFlow-pretrained baseline model. ``SF-ft500" denotes finetuning the SceneFlow-pretrained baseline model with $500$ selected pairs.  ``Ada" denotes the AdaStereo model. ``Ada-ft500" denotes finetuning the AdaStereo model with $500$ selected pairs. ``Scratch" denotes training the baseline model directly on the whole DrivingStereo training set. All models are trained for $100$ epochs. $D1$-error (\%) is adopted for evaluations.}
        \vspace{-2.5mm}
        \resizebox{1\linewidth}{!}{
            \begin{tabular}{ c | c | c | c | c | c  }
            \hline
          Model  & {SF} & {SF-ft500} & {Ada} & {Ada-ft500} & {Scratch}  \\
            \hline
          \adaresnetcorr & 18.0 & 10.2 & 7.8 & 6.3 & 5.4  \\
          \adapsmnet  & 21.4 & 9.1 & 5.2 & 3.9 & 3.2 \\
            \hline
            \end{tabular}
        }
    \label{a2}
    \end{table}
    
    \subsubsection{Verification on More Advanced Models}
    
    In order to further demonstrate the effectiveness of our domain adaptation pipeline, we choose another two advanced $3$-D stereo networks, \emph{i.e.}  GWCNet \cite{guo2019group} and GANet \cite{zhang2019ga}, then extend them as Ada-GWCNet and Ada-GANet respectively. As shown in Tab. \ref{a3}, all of our proposed modules work effectively on top of these more advanced disparity networks. As can be seen, Ada-GWCNet achieves slightly worse cross-domain performance than Ada-PSMNet, while Ada-GANet performs slightly better on two target domains.
    
    \begin{table}[t]
        \centering
        \caption{More verification on top of advanced $3$-D stereo networks. The results on the KITTI and DrivingStereo training sets are reported. $D1$-error (\%) is adopted for evaluations.}
        \vspace{-2.5mm}
        \resizebox{1\linewidth}{!}{
        \begin{tabular}{c | c c c | c | c  }
	        \hline
            \multirow{2}{*}{Model} & {color} & {cost} & {self-supervised} & 
            \multirow{2}{*}{KITTI} & 
            \multirow{2}{*}{DrivingStereo} \\
            {} & {transfer} & {normalization} & {reconstruction} &  {} & {} \\
            \hline
            {\emph{Ada-ResNetCorr}} & {\cmark} & {\cmark} & {\cmark} & 5.0 & 8.0  \\
            \hline
            {\emph{Ada-PSMNet}} & {\cmark} & {\cmark} & {\cmark} & 3.5 & 5.1  \\
            \hline
            \multirow{4}{*}{\emph{Ada-GWCNet}} & 
            {\xmark} & {\xmark} & {\xmark} & 14.9 & 25.8  \\
            {} & {\cmark} & {\xmark} & {\xmark} & 6.0 & 8.3   \\
            {} & {\cmark} & {\cmark} & {\xmark} & 5.2 & 7.1  \\
            {} & {\cmark} & {\cmark} & {\cmark} & 3.7 & 5.5  \\
            \hline
            \multirow{4}{*}{\emph{Ada-GANet}} & 
            {\xmark} & {\xmark} & {\xmark} & 11.3 & 16.3  \\
            {} & {\cmark} & {\xmark} & {\xmark} & 5.0 & 6.7  \\
            {} & {\cmark} & {\cmark} & {\xmark} & 4.4 &  6.0 \\
            {} & {\cmark} & {\cmark} & {\cmark} & 3.3 & 4.8  \\
	    \hline
        \end{tabular}
        }
    \label{a3}
    \end{table}
    
    \subsubsection{New Evaluation Metrics}
    
    Along with the DrivingStereo dataset~\cite{yang2019drivingstereo}, two new metrics are introduced to evaluate the performance of a stereo matching method in driving scenarios: the distance-aware absolute relative difference (ARD) metric, and the semantic-aware matching rate (MR) metric. Specifically, the $ARD_k$ is defined to measure the deviation between disparity prediction $d_p$ and ground-truth $d_g$ in its measurement range as $[k-r, k+r]$, where $k$ denotes the depth, and $r$ denotes the scope. We can draw the ARD curve by linking the $ARD_k$ at different ranges. Moreover, the global difference (GD) of ARD curve is estimated to balance samples among different ranges. The MR metric is defined to compute the matching accuracy on various categories, including vehicle, human, ground, construction, nature, and others. Here, we adopt these two metrics to compare our Ada-PSMNet with the SceneFlow-pretrained PSMNet (baseline) on the KITTI 2015 training set, in which depth and semantic labels are provided. For convenience, we draw the ARD curve and MR radar map to observe the detailed results. As shown in Fig.~\ref{fig-ard_curve}, the ARD curve of Ada-PSMNet is much lower and smoother than that of the SceneFlow-pretrained baseline, indicating better disparity predictions in all ranges from $0m$ to $80m$ after domain adaptation. Moreover, it can be found from the radar map that our Ada-PSMNet substantially outperforms the baseline on all categories of ground, especially vehicle, human, construction, and others. In addition, as shown in Tab. \ref{t16}, the GD of the SceneFlow-pretrained baseline is $20.1\%$, while Ada-PSMNet achieves a GD of $6.3\%$. Hence, the effectiveness and superiority of our domain-adaptive stereo matching pipeline are further demonstrated.
    
        \begin{table}[t]
    \centering
        \caption{Distance-aware ARD \cite{yang2019drivingstereo} comparisons between Ada-PSMNet and the SceneFlow-pretrained PSMNet (baseline) on the KITTI 2015 training set. GD (\%) \cite{yang2019drivingstereo} is adopted for evaluations.}
        \vspace{-2.5mm}
        \resizebox{0.85\linewidth}{!}{
            \begin{tabular}{ c | c c  }
            \hline
          Model  & {GD} & {$D1$-error}  \\
            \hline
          SceneFlow-pretrained PSMNet & 20.1  & 12.1 \\
          \adapsmnet & 6.3 & 3.5 \\
            \hline
            \end{tabular}
        }
    \label{t16}
    \end{table}
    
    \begin{table}[t]
        \centering
        \caption{Comparisons between different source-domain datasets (synthetic SceneFlow,  real-world DrivingStereo). Ada-ResNetCorr and Ada-PSMNet are both adopted for comparisons. $D1$-error (\%) is adopted for evaluations on the KITTI, Middlebury and ETH3D training sets.}
        \vspace{-2.5mm}
        \resizebox{1\linewidth}{!}{
        \begin{tabular}{c | c | c c | c | c  }
	        \hline
            \multirow{2}{*}{Model} &  \multirow{2}{*}{Source} &
            \multicolumn{2}{c|}{KITTI} & 
            {Middlebury} & \multirow{2}{*}{ETH3D} \\
            {} & {} & {2012} & {2015} & {half}  &   \\
            \hline
            \multirow{2}{*}{{\adapsmnet}} & SceneFlow & 3.6 & 3.5 & 8.4  & 4.1 \\
            & DrivingStereo & 2.8 & 3.0 & 8.6  & 3.9 \\
            \hline
            \multirow{2}{*}{{\adaresnetcorr}} & SceneFlow & 5.1 & 5.0 & 12.7  & 5.8 \\
            & DrivingStereo & 4.2 & 4.3 & 13.0  & 5.5 \\
	    \hline
        \end{tabular}
        }
    \label{t14}
    \end{table}
    
    \begin{table}[t]
    \centering
        \caption{Quick adaptation performance on the KITTI and DrivingStereo training sets. The SceneFlow-pretrained PSMNet is adopted as the baseline model. ``Finetune 10'' denotes finetuning the SceneFlow-pretrained baseline model through our AdaStereo pipeline for $10$ epochs. ``From Scratch'' denotes training a domain-adaptive model from scratch through our AdaStereo pipeline for $100$ epochs. $D1$-error (\%) is adopted for evaluations.}
        \vspace{-2.5mm}
        \resizebox{1\linewidth}{!}{
            \begin{tabular}{ c | c | c | c | c  }
            \hline
          Target Domain  & {Baseline} & {Finetune 10} & {Finetune 20} & {From Scratch} \\
            \hline
          KITTI  & 12.1 & 6.4 & 5.5 & 3.5 \\
          DrivingStereo & 20.9 & 8.8 & 7.8 & 5.1   \\
            \hline
            \end{tabular}
        }
    \label{t15}
    \end{table}
    
    
    \subsubsection{Real-world Source Domain}
    
    To further demonstrate the effectiveness and generalization capability of our proposed domain-adaptive stereo matching pipeline, we adopt the real-world DrivingStereo dataset as the source-domain dataset which is large enough for training, then test the domain adaptation performance of Ada-ResNetCorr and Ada-PSMNet on the other three real-world datasets (KITTI, Middleburry and ETH3D). For fair comparisons with domain-adaptive models trained on the SceneFlow dataset, we only use $35k$ training pairs (as many as SceneFlow) uniformly sampled from the DrivingStereo training set, meanwhile other training settings remain unchanged. As can be seen in Tab. \ref{t14}, the error rates of two domain-adaptive models on the KITTI and ETH3D training sets are all reduced, since the outdoor driving scenarios in DrivingStereo are more consistent with these two target domains than SceneFlow. However, the error rates on the Middlebury training set get slightly worse, because of the inconsistent scenarios between DrivingStereo (outdoor) and Middleburry (indoor). To sum up, our method is robust to the choice of source-domain dataset. When there are plenty of real-world stereo pairs with ground-truth disparity annotations, that serve as the source-domain training samples for real-world deployments of domain-adaptive stereo matching systems, reliable disparity predictions can be easily obtained through our domain adaptation pipeline.

    \subsubsection{Quick Adaptations}
    
    Although our method achieves remarkable domain adaptation performance, each domain-adaptive model needs training from scratch for $100$ epochs for each target domain. Hence we conduct additional experiments to verify whether our domain adaptation pipeline can be used for quick adaptations: finetuning a synthetic-data pretrained model through our domain adaptation pipeline on the target domain for a few epochs. In Tab. \ref{t15}, we compare quick adaptations in which the SceneFlow-pretrained PSMNet is finetuned for $10$ or $20$ epochs on the target domain without ground-truths, with training a domain-adaptive model from scratch for 100 epochs. After adaptation with only $20$ epochs, the error rates on two target domains are significantly reduced ($6.6\%$ on KITTI, $13.1\%$ on DrivingStereo), which however, are still slightly inferior to those of training from scratch. To sum up, our proposed domain-adaptive stereo matching pipeline can be easily integrated into quick adaptation application scenarios and real-world deployments, in which reasonable domain adaptation performance can be achieved with only $10\%\sim20\%$ training costs.

\section{Conclusions and Future Work}

In this paper, we address the domain adaptation problem of current deep stereo matching networks. Following the standard domain adaptation methodology, we present a novel domain-adaptive approach called   \textbf{\adastereo} to promote the all-round domain adaptation ability in the stereo matching pipeline. Specifically, three modules are proposed to bridge the domain gaps at different levels: (i) a non-adversarial progressive color transfer algorithm for input-level alignment; (ii) a parameter-free cost normalization layer for internal feature-level alignment; (iii) a highly related self-supervised auxiliary task for output-space alignment. The intensive ablation studies and break-down comparisons on multiple target-domain datasets validate the effectiveness of each module. Benefited from the proposed methodology of  multi-level alignments, our domain-adaptive stereo matching pipeline achieves state-of-the-art cross-domain performance on various target domains, including both indoor and outdoor scenarios. Moreover, our synthetic-data pretrained domain-adaptive models achieve remarkable accuracy on four public stereo matching benchmarks even without finetuning. In the future, we will further develop our AdaStereo to simultaneously adapt to multiple target domains. Besides, we will refine the proposed domain-adaptive pipeline for real-time online adaptations, in which target-domain image pairs are unavailable for training.



\bibliographystyle{spmpsci}      
\bibliography{egbib}

\end{document}